\documentclass[lettersize,journal]{IEEEtran}
\usepackage{amsmath,amsfonts}
\usepackage{algorithmic}
\usepackage{algorithm}
\usepackage{array}
\usepackage[caption=false,font=normalsize,labelfont=sf,textfont=sf]{subfig}
\usepackage{textcomp}
\usepackage{stfloats}
\usepackage{url}
\usepackage{verbatim}
\usepackage{graphicx}
\usepackage{cite}
\usepackage{bbm}
\usepackage{csquotes}
\usepackage{times}
\usepackage{epsfig}
\usepackage{amssymb}
\usepackage{booktabs}
\usepackage[dvipsnames]{xcolor}
\usepackage[normalem]{ulem}
\usepackage[colorlinks=true,linkcolor=black,citecolor=blue,urlcolor=red,]{hyperref}
\usepackage{soul}

\def\ie{\emph{i.e.}}
\def\eg{\emph{e.g.}}

\def\etal{{\em et al.~}}

\begin{document}

\title{Weakly-supervised Contrastive Learning for Unsupervised Object Discovery}


\author{Yunqiu Lv,~
Jing Zhang,~
Nick Barnes,~
Yuchao Dai\\
\IEEEcompsocitemizethanks{\IEEEcompsocthanksitem Yunqiu Lv and Yuchao Dai are with School of Electronics and Information, Northwestern Polytechnical University and Shaanxi Key Laboratory of Information Acquisition and Processing, Xi'an, China.
\IEEEcompsocthanksitem Jing Zhang and Nick Barnes are with School of Computing, Australian National University, Canberra, Australia.
\IEEEcompsocthanksitem
Yuchao Dai (daiyuchao@gmail.com) is the corresponding author. 
\IEEEcompsocthanksitem This research was supported in part by National Natural Science Foundation of China (62271410) and by the Fundamental Research Funds for the Central Universities.
}
}

\markboth{Journal of \LaTeX\ Class Files,~Vol.~14, No.~8, August~2021}%
{Shell \MakeLowercase{\textit{et al.}}: A Sample Article Using IEEEtran.cls for IEEE Journals}


\maketitle

\begin{abstract}

Unsupervised object discovery (UOD) refers to the task of 
discriminating the whole region of objects from the background
within a scene without relying on labeled datasets,
which benefits the task of bounding-box-level localization and pixel-level segmentation.
This task is promising due to its ability to discover objects in a generic manner. We roughly categorise existing techniques into two main directions,
namely the generative solutions based on image resynthesis, and the clustering methods based on self-supervised models.
We have observed that the former heavily relies on the quality of image reconstruction, while the latter shows
limitations in effectively modeling semantic correlations. To directly target at object discovery, we focus on the latter approach and propose a novel solution by incorporating weakly-supervised contrastive learning (WCL) to enhance semantic information exploration. We design a semantic-guided self-supervised learning model to extract high-level semantic features from images, which is achieved
by fine-tuning the feature encoder of 
a self-supervised model, namely DINO, via WCL. 
Subsequently, we introduce Principal Component Analysis (PCA) to localize object regions. The principal projection direction, corresponding to the maximal eigenvalue, serves as an indicator of the object region(s).
Extensive experiments on benchmark unsupervised object discovery datasets demonstrate the effectiveness of our proposed solution. 
The source code and experimental results are publicly available via our project page at \url{https://github.com/npucvr/WSCUOD.git}

\end{abstract}


\section{Introduction}
\label{sec:intro}

\IEEEPARstart{U}{nsupervised} object discovery (UOD)~\cite{chen2019unsupervised,voynov2020unsupervised, object_seg_without_label_icml,locatello2020object, yu2021unsupervised,melas2021finding,large_unsup_obj_dis,caron2021emerging,self_sup_loc_obj_no_labels,Deep_Spectral_Methods_Kyriazi_2022_CVPR,wang2022self,henaff2022object,simeoni2023found,seitzer2023dinosaur,Freesolo2022,ponimatkin2022simple, zhang2020tip_cnn} refers to correctly localizing the semantic-meaningful objects in an image without
any manual annotations. 
It is an important step for modern object detection or segmentation as it can not only reduce the effort of manual annotation in supervised-learning-based methods~\cite{ucnet_sal,carion2020end}, but also offer the pseudo label or initial position prompt for large-scale semi-supervised
methods~\cite{kirillov2023segmentanything}.
Although there exists research on unsupervised semantic segmentation~\cite{groupvit,zadaianchuk2022unsupervised,henaff2022object,sauvalle2022unsupervised,Deep_Spectral_Methods_Kyriazi_2022_CVPR,ke2022unsupervised}, or unsupervised instance segmentation~\cite{Freesolo2022}, we find the former usually relies on extra weak annotations, and the latter is mainly designed on top of unsupervised object discovery models.
In this case, effective unsupervised object discovery models can relieve the labeling requirement,
facilitating many downstream tasks, \eg,~semantic segmentation \cite{FCN_sem_seg}, scene editing \cite{scene_edit}, image retrieval \cite{image_retr_bench}.


\begin{figure}
    \centering
   \begin{center}
   \begin{tabular}{{c@{ } c@{ } c@{ } c@{ }c@{ }}}   {\includegraphics[width=0.180\linewidth,height=0.132\linewidth]{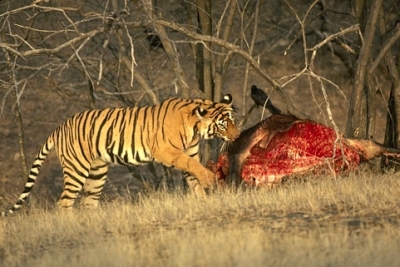}}&   {\includegraphics[width=0.180\linewidth,height=0.132\linewidth]{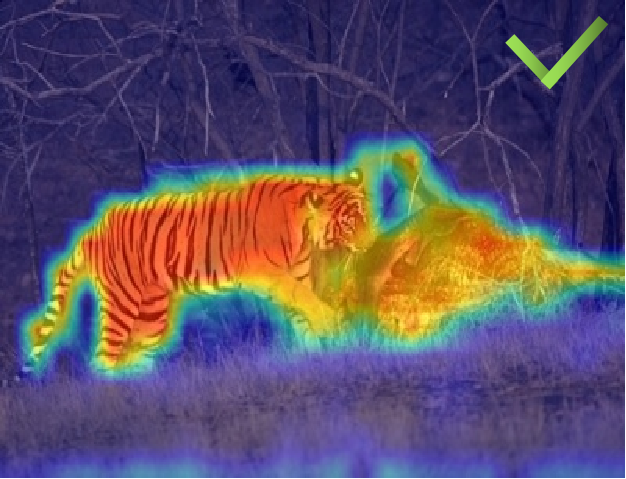}}&    {\includegraphics[width=0.180\linewidth,height=0.132\linewidth]{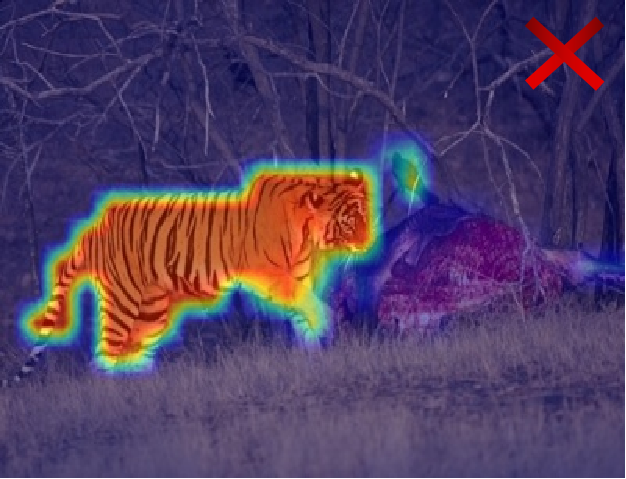}}&    {\includegraphics[width=0.180\linewidth,height=0.132\linewidth]{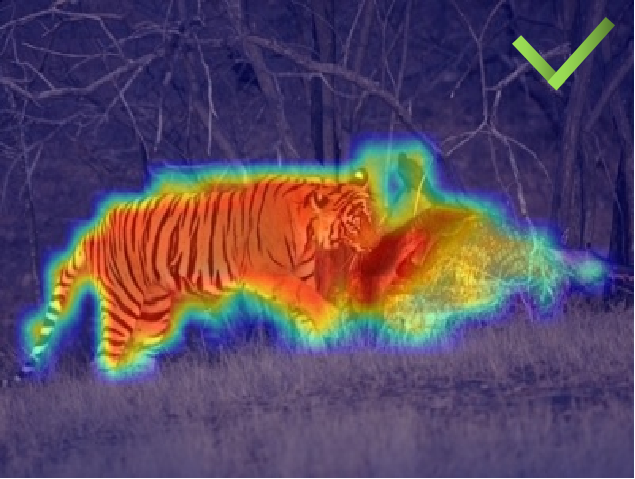}}&    {\includegraphics[width=0.180\linewidth,height=0.132\linewidth]{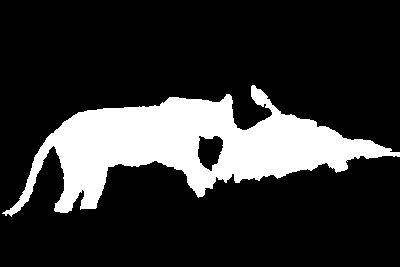}} \\    {\includegraphics[width=0.180\linewidth,height=0.132\linewidth]{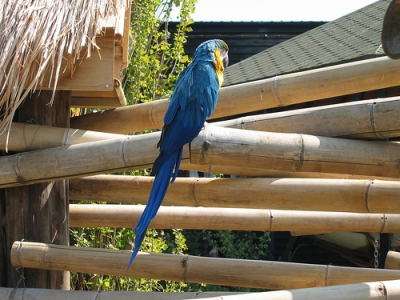}}&    {\includegraphics[width=0.180\linewidth,height=0.132\linewidth]{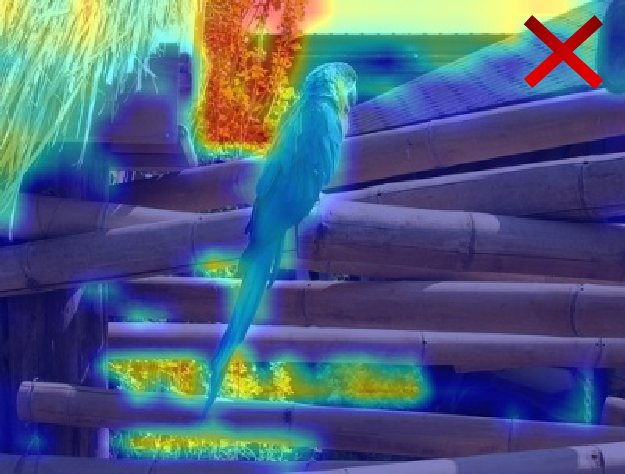}}&     {\includegraphics[width=0.180\linewidth,height=0.132\linewidth]{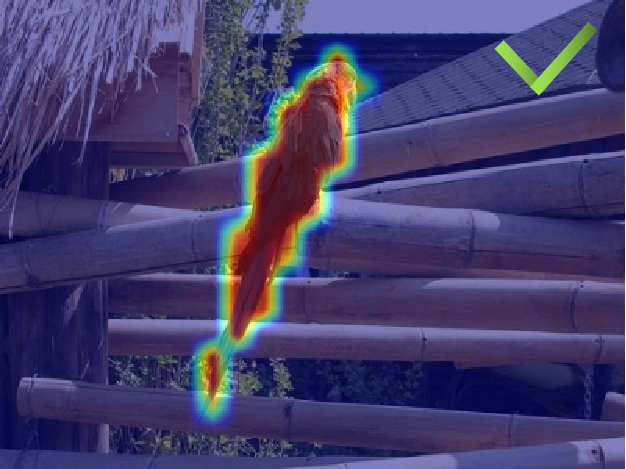}}&    {\includegraphics[width=0.180\linewidth,height=0.132\linewidth]{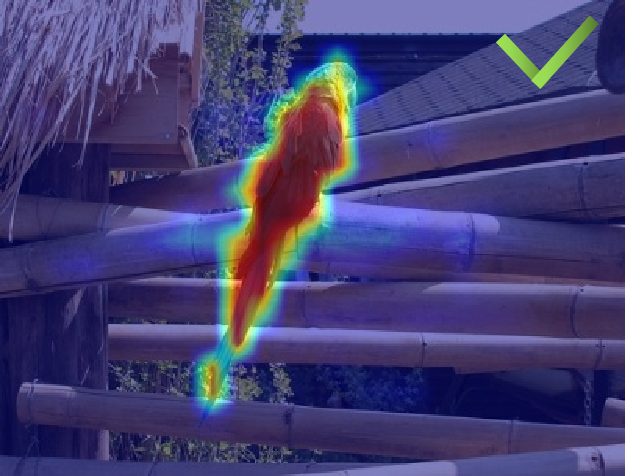}}&    {\includegraphics[width=0.180\linewidth,height=0.132\linewidth]{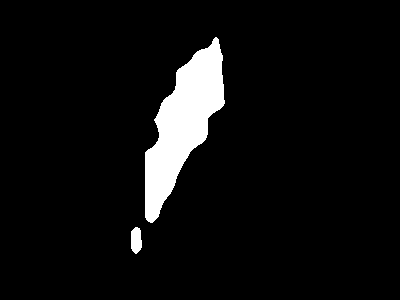}} \\
    \footnotesize{(a) Image} &
    \footnotesize{(b) DINO} &
    \footnotesize{(c) TokenCut} & \footnotesize{(d) Ours} & \footnotesize{(e) GT} \\
   \end{tabular}
   \end{center}
    \caption{
     Object discovery heatmaps of three methods: (a) original RGB images, (b) the principal component heatmap of image representation produced by vanilla DINO~\cite{caron2021emerging}, (c) the result of TokenCut~\cite{wang2022self}, (d) the principal component heatmap of image representation produced by our weakly-supervised contrastive learning based self-supervised learning network, (e) ground truth.
    } 
    \label{fig:overview_results}
\end{figure}%

As no manual annotation is provided, the mainstream of UOD is based on
self-supervised methods, using either generative models via image-resynthesis~\cite{chen2019unsupervised, voynov2020unsupervised, object_seg_without_label_icml,locatello2020object, yu2021unsupervised, melas2021finding} or 
self-supervised representation learning with instance discrimination~\cite{large_unsup_obj_dis, caron2021emerging,self_sup_loc_obj_no_labels,Deep_Spectral_Methods_Kyriazi_2022_CVPR,wang2022self,henaff2022object,xiao2022learning}. We find that the image-synthesis solutions heavily rely  on the generation quality of generative models, and they usually perform unsatisfactorily on
images with complex background. 
The self-supervised model for object discovery, \ie~DINO~\cite{caron2021emerging},
implements instance discrimination/classification~\cite{wu2018unsupervised,grill2020bootstrap}, where augmentations of the same image are regarded as the same class. Then the structures of objects are embedded with feature and attention map of the class
token on the heads of the last layer, which can highlight the salient foreground objects from the image. 
Based on this,
subsequent works \cite{self_sup_loc_obj_no_labels,Deep_Spectral_Methods_Kyriazi_2022_CVPR,wang2022self} are designed to segment the objects using
post-processing methods based on DINO features, such as seed expansion \cite{kolesnikov2016seed} or normalized cut \cite{shi2000normalized}.
However,  
as no semantic guidance is supplied, the attention map 
contains extensive noise for real-world scene-centric images, where the background can be inaccurately activated rather than the foreground.
Further, as only image-level constraints are applied in DINO~\cite{caron2021emerging}, the generated representation focuses more on image-level similarity rather than pixel-level similarity.
which is not ideal for semantic-dominant unsupervised object discovery.

These two main issues, \ie~inaccurate background activation and less informative semantic representation learning, drive us to find a self-supervised representation learning method specifically for unsupervised object discovery, that highlights the semantically meaningful regions of the objects and suppresses background activation.
To achieve this, we incorporate weakly-supervised contrastive learning (WCL)~\cite{zheng2021weakly} into the self-supervised learning framework~\cite{caron2021emerging}, where the involvement of WCL leads to semantic guided feature representation, avoiding the noisy foreground activation in existing solutions~\cite{wang2022self}.

Specifically,
as DINO~\cite{caron2021emerging} is effective in learning
knowledge about objects from numerous object-centric images from ImageNet \cite{russakovsky2015imagenet}, we first employ the ViT model~\cite{dosovitskiy_ViT_ICLR_2021} trained by DINO as our feature extraction encoder. To further enhance the invariant properties
of the object in the representation of scene-centric 
images, we fine-tune the self-supervised model
with weakly-supervised contrastive loss~\cite{zheng2021weakly}
on a scene-centric dataset without manual annotations, \ie~images from the DUTS \cite{imagesaliency} training dataset. 
The naive weakly-supervised contrastive loss is designed based on image-level
positive/negative pairs
to encourage the object-centric representation 
of images.
In our dense prediction setting, we
design an alignment loss to 
enforce the pixel-level semantic coherence in the overlapping part of different views.


With the proposed weakly-supervised contrastive learning, the intrinsic semantic information is embedded and becomes prominent in the image representation, while the background is suppressed.
Although the foreground representation can still vary, the suppressed background representation becomes similar with lower variance. We then perform principal component analysis (PCA)~\cite{pca1901} to extract the principal component of the image representation as the descriptor of the object, and localize
the object by finding the pixels with a high correlation with the object descriptor. 

In summary, our contributions are threefold. 
\begin{itemize}
    \item We extend weakly-supervised contrastive learning to unsupervised object discovery for the first time and introduce an alignment loss to keep the pixel-level semantic consistency for object segmentation. 
    \item We propose an effective method to extract the most discriminative region of the image representation via principal component analysis.
    \item Experiments on unsupervised object segmentation, unsupervised object detection and video object detection tasks demonstrate the effectiveness of our proposed solution. 
\end{itemize}
    





\section{Related Work}
\subsection{Unsupervised Object Discovery}

Unsupervised object discovery aims to generate class-agnostic bounding boxes or segmentation maps to capture semantically-meaningful objects
to be used in downstream vision tasks.
As 
pointed out in~\cite{co_attention_cnn_uns_co_seg,wei2019unsupervised,unsup_multi_obj_dis},
the ability of deep neural networks heavily depends on 
large-scale labeled
datasets, 
and reducing the dependency has been 
tackled from
two directions, \ie~generative methods via image-resynthesis, and self-supervised representation learning.


Generative models~\cite{vae_bayes_kumar,pang2020learning,GAN_nips} are achieved via image-resynthesis,
\ie~\cite{chen2019unsupervised,abdal2021labels4free,locatello2020object,singh2021illiterate, yu2021unsupervised, seitzer2023dinosaur}, 
introduce
latent codes
for object and background
respectively, and then recover the image collaboratively with an energy-based model~\cite{pang2020learning} or generative adversarial nets (GAN)~\cite{GAN_nips}. 
Other methods~\cite{Object_Segmentation_Without_Labels_with_Large-Scale_Generative_Models, voynov2020unsupervised, melas2021finding,object_seg_without_label_icml} aim
to find an interpretable direction representing the semantic meaning of \enquote{saliency lighting} or \enquote{background removing}
in the latent space of a GAN and obtain object segmentation by adjusting along the latent direction. However, the dependency on reconstruction quality makes object discovery difficult when the training dataset has complex layout and multiple object categories.

Self-supervised representation learning based on instance discrimination
aims to preserve the discriminative features of the object, and discover the object based on an objectness prior. DINO~\cite{caron2021emerging} trained on ViT \cite{dosovitskiy_ViT_ICLR_2021} has shown superior performance on object segmentation even compared with supervised methods. Then, LOST \cite{self_sup_loc_obj_no_labels} and TokenCut \cite{wang2022self} explore the direct usage of DINO features on segmentation through seed expansion and normalized cut respectively. LOST \cite{self_sup_loc_obj_no_labels} localizes the object following the assumption that patches in an object correlate positively with each other and negatively with the background, and the foreground covers less area. 
TokenCut~\cite{wang2022self} uses spectral clustering on the DINO feature and selects the region with maximal eigenvalue as the object. Xie \etal \cite{xie2021unsupervised} employ image retrieval and design both image-level and object-level contrastive learning to encourage the semantic consistency 
in similar objects. Vo \etal \cite{large_unsup_obj_dis} use self-supervised features and obtain the object through PageRank on a proposal-based graph. Hénaff \etal \cite{henaff2022object} propose to simultaneously optimize object discovery and self-supervised representation networks and these two tasks benefit from each other.
With 
pre-trained self-supervised model, other methods use coarse masks generated from the affinity matrix of the features from the pre-trained models as the pseudo label to further train the segmentation/detection model~\cite{Freesolo2022, shin2022unsupervised, simeoni2023found, wang2023maskcut}, or to further solve the graph optimization problem~\cite{ponimatkin2022simple}.
In our setting, we also use the DINO pre-trained model, but we have developed an object-centric representation learning method based on weakly-supervised contrastive learning to prevent the interruption from complex background. 



\subsection{Self-supervised Object-Centric Learning}
The goal of self-supervised object-centric learning is to learn task-agnostic image inherent features. In this paper, we mainly discuss instance discrimination, which 
learns a distinguished representation from data augmentations by considering the augmented views generated from the same image as the same class.
BYOL \cite{grill2020bootstrap} learns a high-quality visual representation by matching the prediction of a network to the output of a different view produced by a target network. DINO \cite{caron2021emerging} extends the work with a different loss and employs ViT as the network architecture. SimSiam \cite{chen2021exploring} improves BYOL with contrastive learning in a Siamese structure. Since naive pairwise feature comparison induces class collapse, many approaches simultaneously cluster the features while imposing
intra-cluster consistency. SwAV \cite{caron2020unsupervised} assigns features to prototype codes and allows codes of two views to supervise each other. Wen \etal \cite{wen2022self} extend the prototype idea to pixel level. 
WCL \cite{zheng2021weakly} also uses the clustering technique to generate semantic components and performs contrastive learning on different components. WCL has a simpler setting but comparable performance with prototype-based models. Since the object usually has the highest information density in an image, we argue that self-supervised representations can be further explored
in object discovery.

\subsection{Unsupervised Saliency Detection}
Unsupervised saliency detection indicates methods that find and segment the generic objects that draw human attention most quickly, without any supervision from a large-scale dataset. Early works \cite{global_con_mm,saliency_drfi,saliency_sparse_coding,geodesic_saliency,yan2013hierarchical,Manifold-Ranking:CVPR-2013,saliency_optimization_rbd} use hand-crafted features and localize the salient objects based on saliency priors such as contrast prior \cite{itti1998model}, center prior \cite{judd2009learning} or boundary prior \cite{geodesic_saliency}.  The modern literature \cite{croitoru2019unsupervised,GaoSPWK21,jing_cvpr18_noisy_saliency,deepups_saliency,zhang2020learning_eccv} uses a pseudo saliency map in the training of a deep segmentation network. SBF \cite{GaoSPWK21} proposes to train a network by using the pseudo labels fused by multiple traditional methods on super-pixel and image levels. USD \cite{jing_cvpr18_noisy_saliency} designs a noise prediction module to eliminate the effect of noise caused by multiple hand-crafted pseudo masks in training. This idea was extended by employing alternating back-propagation in \cite{zhang2020learning_eccv} to learn the saliency map only from a single pseudo mask. However, these methods still rely on a backbone model pretrained on the labeled ImageNet dataset. SelfMask \cite{shin2022unsupervised} leverages self-supervised features from MoCo~\cite{moco_he_2020}, SwAW~\cite{caron2020unsupervised} and DINO~\cite{caron2021emerging} and selects the salient masks based on object priors by a voting mechanism. In the following, we show that unsupervised object discovery can provide an effective baseline for unsupervised saliency detection.

\subsection{Uniqueness of Our Solution}

Given the pre-trained self-supervised model, such as the ViT model trained by DINO~\cite{caron2021emerging}, existing methods~\cite{self_sup_loc_obj_no_labels, wang2022self} discover the objects by using the compactness property of the foreground object to localize the object and employ the affinity matrix to explore the whole region of the object. However, the performance of these methods is restricted to the performance of DINO pre-trained model, making them vulnerable to complex background. The direct method to address this problem is finetuning the model with scene-centric images. However, due to the noises introduced by the cluttered background, the performance on finetuned DINO-based model becomes even worse according to our experiment in Table~\ref{tab:ablation_study_cod}. The uniqueness of our method is that we have designed a more effective finetuning strategy based on weakly-supervised contrastive learning~\cite{zheng2021weakly} to suppress the activation of the background. The proposed pixel-level alignment loss inspired by~\cite{xiao2022learning} can keep the pixel-level consistency of the object by using images from a scene-centric dataset~\cite{imagesaliency}.
Further, our method has a simple pipeline, consisting of finetuning the ViT pre-trained model with WCL and PCA on the learned features.
Different from other methods that reply on training for unsupervised segmentation/detection~\cite{simeoni2023found,seitzer2023dinosaur,Freesolo2022}, our method does not need any pseudo segmentation or bounding box annotations, making it convenient for our method to generalize to these methods by functioning as the pseudo label for downstream tasks.

\begin{figure}
    \centering
   \begin{center}
   \begin{tabular}{{c@{ }}}
  {\includegraphics[width=0.95\linewidth]{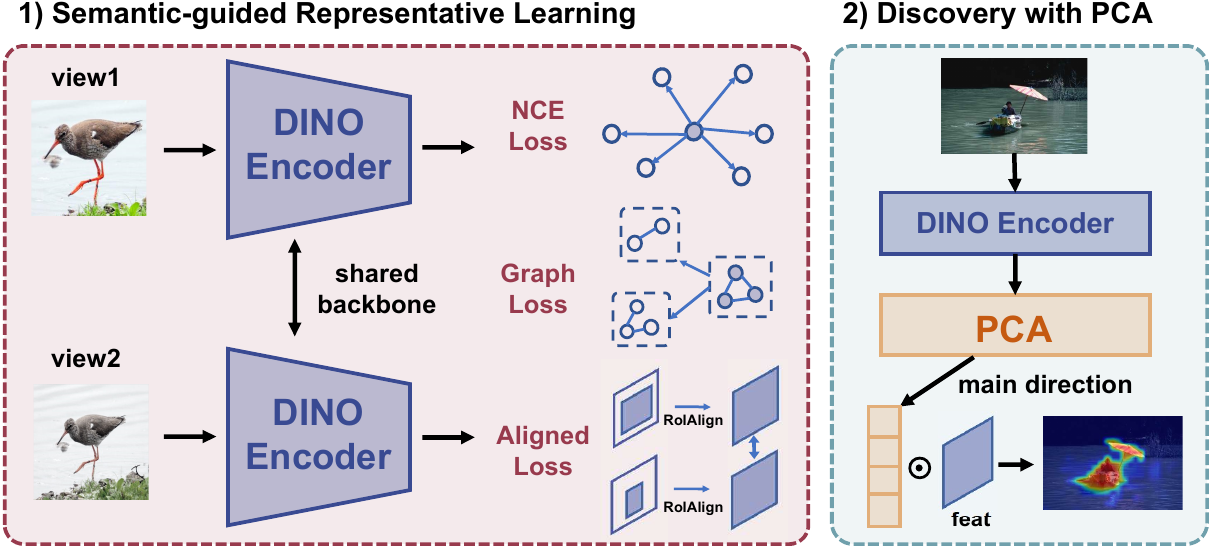}}\\
   \end{tabular}
   \end{center}
    \caption{Overview of the proposed semantic-guided representation learning based unsupervised object discovery model, where the weakly-supervised contrastive learning with pixel-level alignment loss is presented to achieve semantic-guided representation learning, and PCA is then adopt to extract the foreground pixels with a high correlation with the principle object descriptor.
    } 
    \label{fig:overview_method}
\end{figure}%

\section{Our Method}
We work on unsupervised object discovery~\cite{vo2019unsupervised, co_attention_cnn_uns_co_seg,wei2019unsupervised,unsup_multi_obj_dis,Object_Segmentation_Without_Labels_with_Large-Scale_Generative_Models, voynov2020unsupervised, melas2021finding}, aiming to localize generic object(s) of the scene. Without access to ground truth annotations, we rely on self-supervised DINO models to roughly identify the candidate object(s) 
(Sec.~\ref{sec:self_sup_rep_learning}). Due to the absence of high-level semantic information modeling, we find the transformer features trained by DINO
fail to encode accurate semantic information for unsupervised object discovery. We then introduce weakly-supervised contrastive learning~\cite{zheng2021weakly} to the self-supervised learning framework with pixel-level alignment loss to learn a semantic dominated
image representation with rich pixel-level semantic correlation modeling
(Sec.~\ref{sec:rep_learning_wcl}). 
Based on the learned feature, principal component analysis is introduced to generate the pixel-wise object segmentation map (Sec.~\ref{sec:object_region_determine}). An overview of our proposed method is shown in Fig.~\ref{fig:overview_method}.

\subsection{Self-supervised Representation Learning}
\label{sec:self_sup_rep_learning}
Recent work based on self-supervised transformers, \ie~DINO~\cite{caron2021emerging}, has shown impressive performance on the task of unsupervised object discovery. 
DINO trains a
visual transformer
\cite{dosovitskiy_ViT_ICLR_2021} in a
self-supervised manner
to generate the image representation. The model is formulated with a teacher-student network that accepts two views of the same image as input and urges the output of the student network with parameters $\theta_s$ to match that of the teacher network with parameters $\theta_t$. During training, parameters of the student network ($\theta_s$) are updated via 
the loss function: 
\begin{equation}
    \mathcal{L}_{\text{dino}}=\frac{1}{2}(\text{CE}(P_t(\mathbf{x}), P_s(\mathbf{x}') + \text{CE}(P_t(\mathbf{x}'), P_s(\mathbf{x})),
\end{equation}
where $P_t$ and $P_s$ indicate the Softmax outputs 
generated by teacher and student network respectively. $\mathbf{x}$ and $\mathbf{x}'$ are different views of the same image. $\text{CE}(\cdot,\cdot)$ computes the cross-entropy loss.
Parameters of the teacher network are obtained by an exponential moving average (EMA) on the student weights, \ie~a momentum encoder~\cite{moco_he_2020}, leading to: $\theta_t\gets\lambda\theta_t+(1-\lambda)\theta_s$, where $\lambda$ is designed to follow a cosine schedule from 0.996 to 1 during training~\cite{grill2020bootstrap}.

DINO has demonstrated that
self-supervised
ViT features~\cite{dosovitskiy_ViT_ICLR_2021}
contain more explicit scene layout and object boundaries compared with supervised training.
Thresholding its attention maps 
based on the \texttt{[CLS]}
token of the last layer can also generate a high-quality object segmentation map. Further, methods such as LOST~\cite{self_sup_loc_obj_no_labels} and TokenCut~\cite{wang2022self} that operate directly on the vanilla DINO features have achieved favorable results on simple images for object discovery.
However, as both methods assume DINO features are reliable in representing generic object(s), they fail to localize objects in complex scenarios when DINO features are no longer reliable.


As it is pretrained on object-centric images in ImageNet,
the DINO representation is usually
noisy especially in natural scene-centric images.
Additionally, the training only matches different views of the same image without considering inter-image relationships,
making it vulnerable to the complex components in scene-centric images. We then resort to contrastive learning~\cite{chopra2005learning,dimension_reduction_lecun} to learn inter-image correlation for effective image representation learning.

\subsection{Semantic-guided Representation Learning}
\label{sec:rep_learning_wcl}
We introduce weakly-supervised contrastive learning to our unsupervised object discovery task to achieve semantic-guided representation learning.

\noindent\textbf{Contrastive learning:} Contrastive loss~\cite{chopra2005learning,dimension_reduction_lecun} was introduced for metric learning, which takes pair of examples ($\mathbf{x}$ and $\mathbf{x}'$) as input and trains the network ($f$) to predict whether they are similar (from the same class: $\mathbf{y}_\mathbf{x}=\mathbf{y}_{\mathbf{x}'}$) or dissimilar (from different classes: $\mathbf{y}_\mathbf{x}\neq\mathbf{y}_{\mathbf{x}'}$).
Contrastive learning 
has self-supervised or unsupervised 
approaches, aiming to learn a feature representation $f_\mathbf{x}=f_\theta(\mathbf{x})$ for an image $\mathbf{x}$, where $f_\mathbf{x}$ can represent $\mathbf{x}$ in low-dimension space for downstream tasks,~\ie~image classification and semantic segmentation. To achieve self-supervised learning,
positive/negative pairs are constructed via data augmentation techniques~\cite{xie2021propagate,li2021dense,wang2021dense,van2021unsupervised,o2020unsupervised,chaitanya2020contrastive,xie2021detco}, where the basic principle is that similar concepts should have similar representations, and thus stay close to each other in the embedding space. On the contrary, dissimilar concepts should stay apart in the embedding space.

Taking a step further, triplet loss~\cite{Distance_Metric_Learning,large_scale_online_learning,facenet} achieves metric learning by using triplets, including a query sample ($\mathbf{x}$), its positive sample ($\mathbf{x}^{+}$) and negative sample ($\mathbf{x}^{-}$). In this case, the goal of triplet loss is to minimize the distance between $\mathbf{x}$ and its positive sample $\mathbf{x}^{+}$, and maximize the distance between $\mathbf{x}$ and its negative sample $\mathbf{x}^{-}$.
A key issue for triplet loss is that it only learns from one negative sample, ignoring the dissimilarity with all the other candidate negative samples, leading to unbalanced metric learning. To solve this problem, \cite{npair_loss} introduces an $N$-pair loss to learn from multiple negative samples.
Similarly, InfoNCE loss~\cite{oord2018representation} uses categorical cross-entropy loss to identify positive samples from a set of negative noise samples~\cite{chopra2005learning,Noise_contrastive_estimation_aistats2010}, which optimizes the negative log-probability of classifying the positive sample correctly via:
\begin{equation}
\label{infornce_loss}
\mathcal{L}_{N C E}=-\log \frac{\exp \left(\operatorname{sim}\left(\mathbf{z}_i, \mathbf{z}_j\right) / \tau\right)}{\sum_{k=1}^N \mathbbm{1}_{[k \neq i]} \exp \left(\operatorname{sim}\left(\mathbf{z}_i, \mathbf{z}_k\right) / \tau\right)},
\end{equation}
where $(\mathbf{z}_i, \mathbf{z}_j)$ is the positive pair whose elements are the feature embeddings of image augmentation views from the same image, while $(\mathbf{z}_i, \mathbf{z}_k)$ are either the positive pair or the negative pair, $\operatorname{sim}(\mathbf{z}_i, \mathbf{z}_j)=\mathbf{z}_i^T\mathbf{z}_j/\|\mathbf{z}_i\|\|\mathbf{z}_j\|$ is the dot product between the $\ell_2$ normalized feature embedding of $\mathbf{z}_i$ and $\mathbf{z}_j$. Each embedding is produced by a projection head after the encoder network.

\noindent\textbf{Weakly-supervised contrastive learning:} As contrastive learning treats all negative samples equally, class collision can occur where the model simply pushes away all negative pairs (which may be positive semantically) with equal degree.
The main motivation of weakly-supervised contrastive learning~\cite{zheng2021weakly} is to avoid the class collision problem.
Specifically, a two-projection heads based framework following SimCLR~\cite{chen2020simple} was introduced to achieve semantic guided representation learning.
Then,  
weakly-supervised contrastive learning~\cite{zheng2021weakly} assigns the same weak labels to samples that share similar semantics and uses contrastive learning to push apart the negative pairs with different labels. The label assignment is obtained by a connected component labeling (CCL) process. Specifically, a 1-nearest neighbour graph $G=(V,E)$ is constructed from a batch of samples. Each vertex is the feature embedding of each image, which is produced by a projection head with DINO feature as input. The edge $e_{ij}=1$ if sample $i$ is the 1-nearest neighbor of sample $j$ and vice versa. Graph segmentation is then achieved by
the the Hoshen-Kopelman algorithm~\cite{hoshen1976percolation}. With the CCL process, samples of the same component after graph segmentation are given the same weak label.

For implementation, in addition to the base encoder $f(\cdot)$ and projection head $g(\cdot)$ used by
SimCLR~\cite{chen2020simple},
\cite{zheng2021weakly} introduces an auxiliary projection head $\phi(\cdot)$ which shares the same structure as $g(\cdot)$. The goal of $\phi(\cdot)$ is to explore similar samples across the images and generate weak labels $y\in\mathbb{R}^{N\times N}$ as a supervisory signal to attract similar samples, where $N$ is the size of the minibatch, $y_{i,j}=1$ means $x_i$ and $x_j$ are similar, and $y_{i,j}=0$ indicates 
$x_i$ and $x_j$ are a negative pair.
Given the weak label $y\in\mathbb{R}^{N\times N}$, \cite{zheng2021weakly} achieves supervised contrastive loss 
using: 
$\mathcal{L}_{sup}=\frac{1}{N}\sum_{i=1}^N\mathcal{L}^i$, where $\mathcal{L}^i$ is defined as:
\begin{equation}
    \label{weak_label_supervised_contrastive}
    \mathcal{L}^i=-\sum_{j=1}^N \mathbbm{1}_{\mathbf{y}_{i,j}=1} \log \frac{\exp \left(\operatorname{sim}\left(\mathbf{v}_i, \mathbf{v}_j\right) / \tau\right)}{\sum_{k=1}^N \mathbbm{1}_{[k \neq i]} \exp \left(\operatorname{sim}\left(\mathbf{v}_i, \mathbf{v}_k\right) / \tau\right)}.
\end{equation}
$(\mathbf{v}_i,\mathbf{v}_j)$ are the embeddings of the positive pair within the same semantic group, generated by the projection head $\phi(\cdot)$.
$\tau$
regularizes the sharpness of the output distribution.

\textit{Similarity regularization:} With the weakly-supervised contrastive loss function in Eq.~\eqref{weak_label_supervised_contrastive}, similar samples, \ie~$\mathbf{v}_i$ and $\mathbf{v}_j$, should have similar representations. To avoid error propagation due to less accurate graph segmentation, \enquote{similarity regularization} is applied to guarantee that the similarity between $\mathbf{v}_i$ and $\mathbf{v}_j$ is not extremely high. Particularly, we perform
two types of augmentations for samples in the mini-batch, named $V_1$ and $V_2$, 
and the semantic groups are swapped to supervise each other. 
Consequently, we define the graph loss $\mathcal{L}_{\text {graph }}$ as:
\begin{equation}
\label{graph_loss}
\mathcal{L}_{\text {graph }}=\mathcal{L}_{\text {sup }}\left(V^1, \mathbf{y}^2\right)+\mathcal{L}_{\text {sup }}\left(V^2, \mathbf{y}^1\right),
\end{equation}
where $\mathbf{y}^1$ and $\mathbf{y}^2$ are the weak labels of $V^1$ and $V^2$, respectively, and $\mathcal{L}_{\text {sup }}$ computes the weakly-supervised contrastive loss as shown in Eq.~\eqref{weak_label_supervised_contrastive}, \ie~$\mathcal{L}_{\text {sup }}=\frac{1}{B}\sum_{i=1}^B\mathcal{L}^i$, with $B$ as batch size.

\noindent\textbf{Semantic enhanced representation learning:}
To enhance semantic-level correlation for effective representation learning, we finetune a pre-trained DINO~\cite{caron2021emerging} model with an object-centric dataset, \ie~images from DUTS~\cite{imagesaliency} dataset, via weakly-supervised contrastive learning.

There are two main components in the weakly-supervised contrastive learning framework: the vanilla contrastive loss as shown in Eq.~\eqref{infornce_loss},
that targets at learning instance discrimination; and
the graph loss in Eq.~\eqref{graph_loss}, aiming to
find common features among images sharing similar semantic information.
As the vanilla contrastive loss and weakly-supervised contrastive loss only measure the similarity at image-level, the pixel-level semantic alignment loss is designed to further explore the object region by keeping the pixel-level consistency of the same object across different views of the same image. Assuming $\mathbf{x}_i$ and $\mathbf{x}_j$ are different views generated by applying data augmentation on the same image, we apply the backbone network $f$ and the projection head $\eta$ to obtain their features $\mathbf{s}_i=\eta(f(\mathbf{x}_i^o))$ and $\mathbf{s}_j=\eta(f(\mathbf{x}_j^o))$. Then $\mathbf{x}_i$ and $\mathbf{x}_j$ are projected into the original image by removing the data augmentation in order to obtain their overlapping regions and the corresponding features are cropped from $\mathbf{s}_i$ and $\mathbf{s}_j$. After that, ROIAlign \cite{he2017mask} is used to align the features to the same scale. Therefore, we can get $\mathbf{s}_i^o=\operatorname{ROIAlign}(\operatorname{CROP}(\mathbf{s}_i))$ and $\mathbf{s}_j^o=\operatorname{ROIAlign}(\operatorname{CROP}(\mathbf{s}_j))$. For simplicity, we eliminate the superscript and denote them as $\mathbf{s}_i$ and $\mathbf{s}_j$.
The alignment loss is then defined as:
\begin{equation}
\label{eq:alignment_loss}
\begin{aligned}
    \mathcal{L}_{\text{align}}=\frac{1}{chw}\sum_c\sum_{p,q}\left({\text{CE}}(\mathbf{s}_i^{(p,q)}, \mathbf{s}_j^{(p,q)})
    +{\text{CE}}(\mathbf{s}_j^{(p,q)}, \mathbf{s}_i^{(p,q)})\right),
\end{aligned}
\end{equation}
where $(p,q)$ is the coordinate, $\mathbf{s}_i^{(p,q)}\in\mathbb{R}^c$ is the feature vector on position $(p,q)$, indicating pixel-to-pixel alignment, $c,h,w$ indicate channel dimension, height and width of the feature $\mathbf{s}$, respectively.
With both the weakly-supervised contrastive loss to explore inter-image semantic correlations, and alignment loss for object-emphasized representation learning, we obtain our final objective function as:
\begin{equation}
\mathcal{L}=\mathcal{L}_{NCE}+\alpha\mathcal{L}_{\text{graph}}+\beta\mathcal{L}_{\text{align}},
\end{equation}
where $\alpha$ and $\beta$ are set empirically as $\alpha=5$ and $\beta=1$.


\begin{figure}
    \centering
   \begin{center}
   \begin{tabular}{{c@{ } c@{ } c@{ } c@{ }}}   
   {\includegraphics[width=0.23\linewidth]{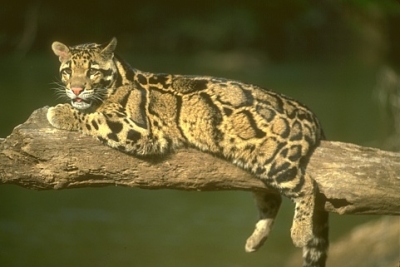}}& 
   {\includegraphics[width=0.23\linewidth]{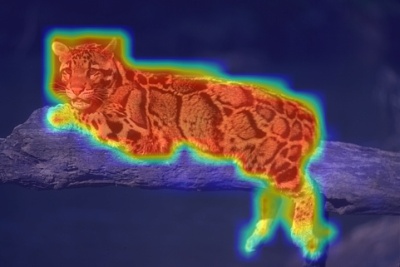}}&   
   {\includegraphics[width=0.23\linewidth]{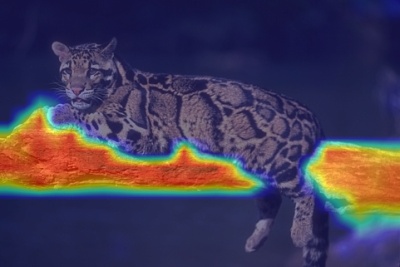}}&    
   {\includegraphics[width=0.23\linewidth]{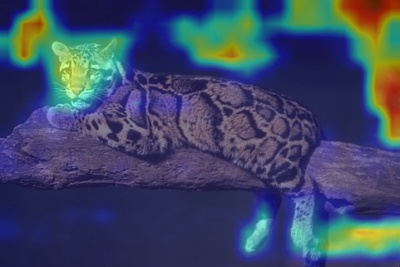}}\\
   {\includegraphics[width=0.23\linewidth]{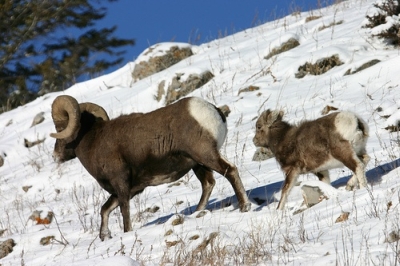}}& 
   {\includegraphics[width=0.23\linewidth]{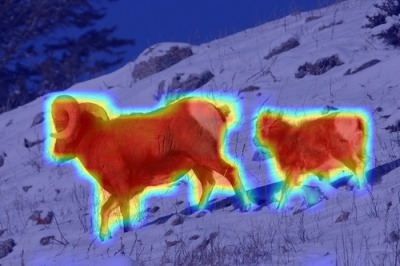}}&   
   {\includegraphics[width=0.23\linewidth]{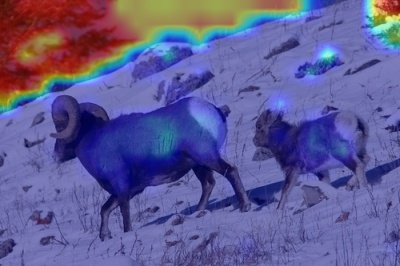}}&    
   {\includegraphics[width=0.23\linewidth]{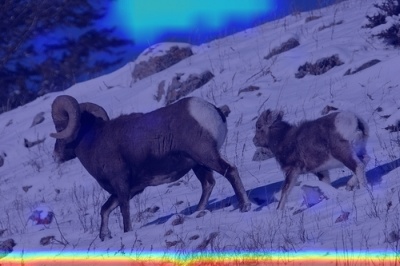}}\\
    \footnotesize{(a) Image} &
    \footnotesize{(b) $1^{st}$ eigvec} &
    \footnotesize{(c) $2^{nd}$ eigvec} &
    \footnotesize{(d) $3^{rd}$ eigvec} \\
   \end{tabular}
   \end{center}
    \caption{The projection maps of the top 3 principal components, where the first principal component correlates closely with the foreground object(s).
    } 
    \label{fig:pca_eigenvectors}
\end{figure}%

\begin{figure*}[tp]
   \begin{center}
   \begin{tabular}{{c@{ } c@{ } c@{ } c@{ } c@{ } c@{ }}}
    {\includegraphics[width=0.155\linewidth]{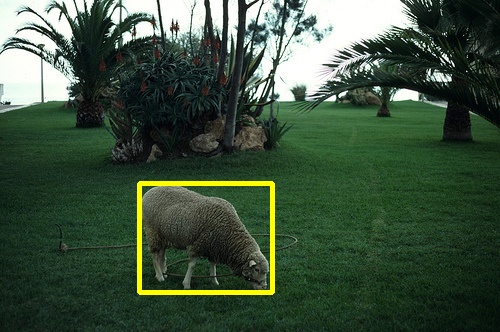}}&
    {\includegraphics[width=0.155\linewidth]{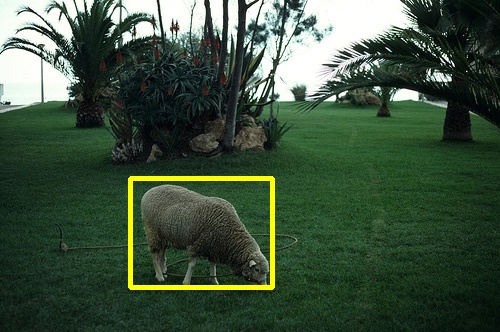}}&
    {\includegraphics[width=0.155\linewidth]{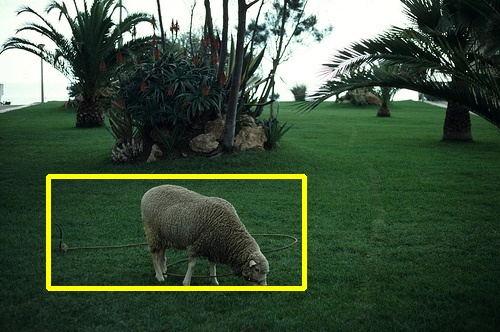}}&
    {\includegraphics[width=0.155\linewidth]{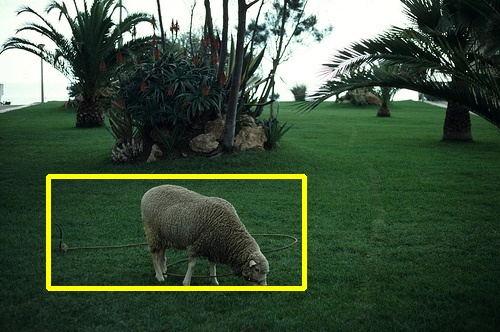}}&
    {\includegraphics[width=0.155\linewidth]{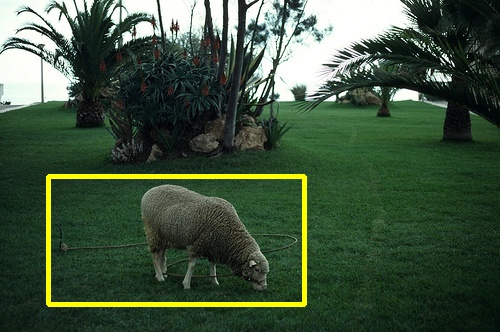}}&
    {\includegraphics[width=0.155\linewidth]{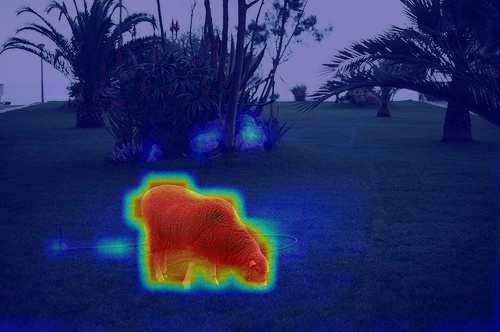}}\\
    {\includegraphics[width=0.155\linewidth]{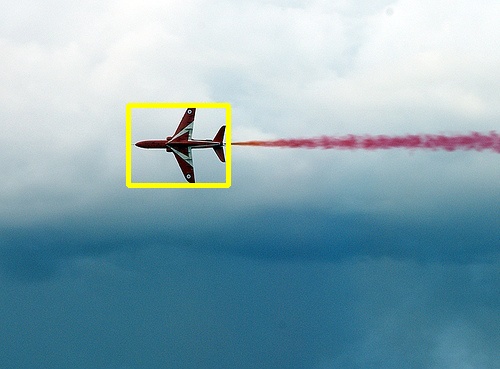}}&
    {\includegraphics[width=0.155\linewidth]{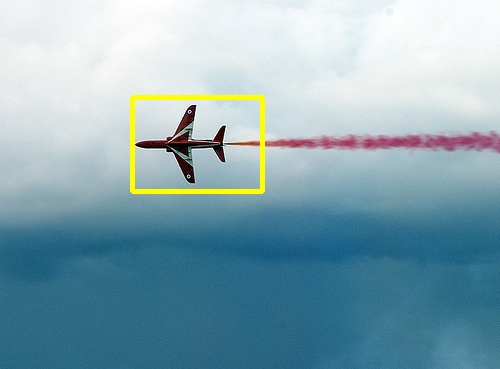}}&
    {\includegraphics[width=0.155\linewidth]{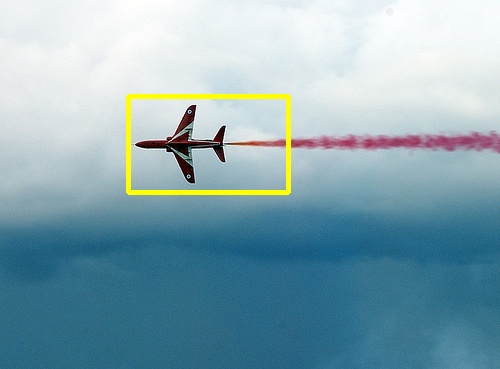}}&
    {\includegraphics[width=0.155\linewidth]{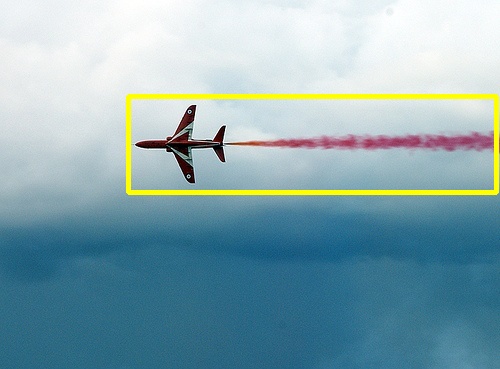}}&
    {\includegraphics[width=0.155\linewidth]{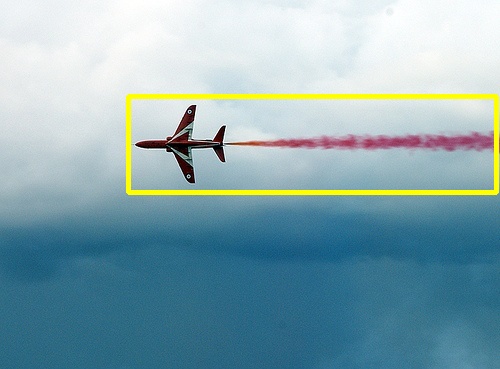}}&
    {\includegraphics[width=0.155\linewidth]{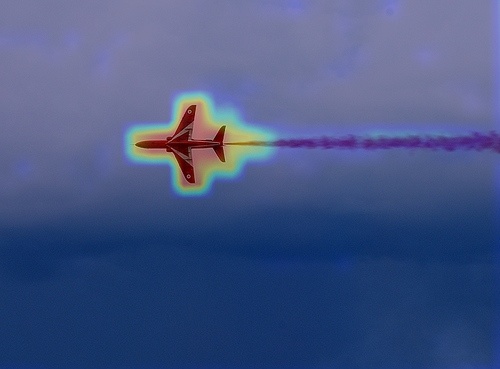}}\\
    {\includegraphics[width=0.155\linewidth]{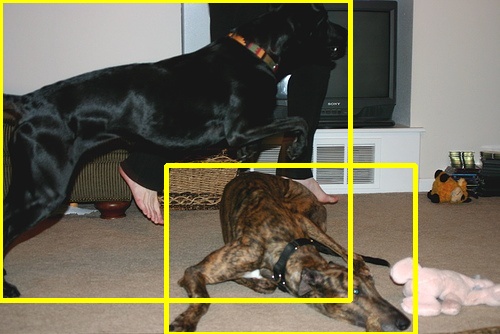}}&
    {\includegraphics[width=0.155\linewidth]{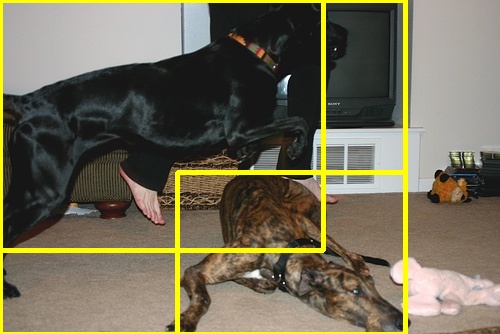}}&
    {\includegraphics[width=0.155\linewidth]{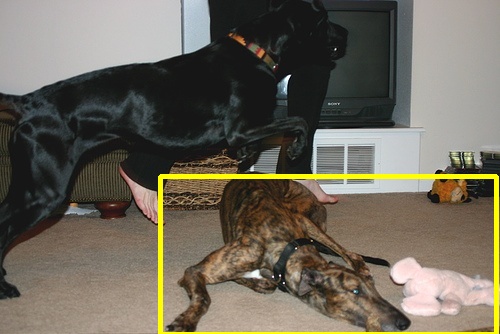}}&
    {\includegraphics[width=0.155\linewidth]{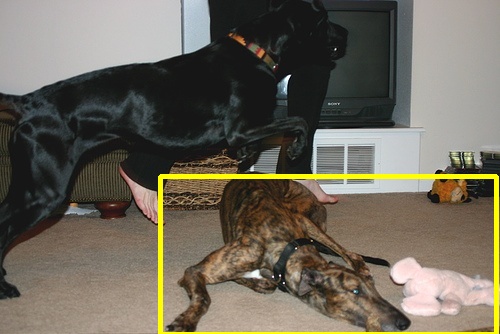}}&
    {\includegraphics[width=0.155\linewidth]{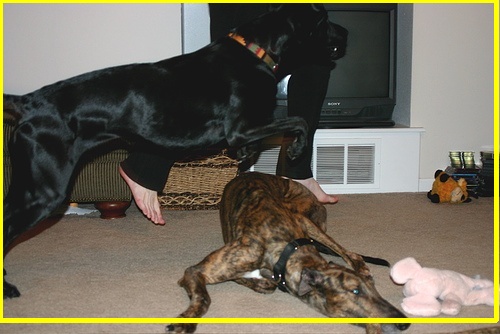}}&
    {\includegraphics[width=0.155\linewidth]{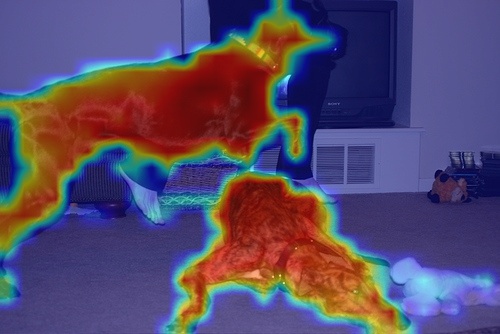}}\\
     \footnotesize{GT} &\footnotesize{Ours}  &\footnotesize{LOST\cite{self_sup_loc_obj_no_labels}} &\footnotesize{TokenCut \cite{wang2022self}} &\footnotesize{DeepSpectral\cite{Deep_Spectral_Methods_Kyriazi_2022_CVPR}}&\footnotesize{Ours$\_$Heatmap} \\
   \end{tabular}
   \end{center}
    \caption{Visual comparison of our model with the existing techniques on object detection, where we also show our generated heat maps (\enquote{Ours\_Heatmap}), based on which our bounding box based object detection results are generated (\enquote{Ours}).
    } 
    \label{fig:obj_detect_comparison}
\end{figure*}

\subsection{Object Discovery based on PCA}
\label{sec:object_region_determine}
With the proposed semantic-guided weakly-supervised contrastive learning, the intrinsic semantic information is embedded and becomes prominent in the image representation, while the background is suppressed.
We further introduce principal component analysis (PCA)~\cite{pca1901}
to our framework to extract object regions for our unsupervised object discovery task.

As illustrated in the bottom-right of Fig.~\ref{fig:overview_results}, the principal descriptor of the image obtained from the vanilla DINO feature~\cite{caron2021emerging} 
contains too much noise from the image background, which is undesirable for object discovery. Although existing solutions~\cite{wang2022self, Deep_Spectral_Methods_Kyriazi_2022_CVPR} can segment the foreground based on clustering techniques, \eg~spectral clustering,
their dependence on the similarity matrix leads to their inability to discover 
heterogeneous foreground objects, as can be seen at the top of the second column in Fig.~\ref{fig:overview_results}.
Instead of relying on computing  pixel-level/patch-level similarity,
we use PCA to find the main descriptor of the image for
object discovery. A similar idea is also applied in deep descriptor transformation (DDT)~\cite{wei2019unsupervised}. However, given a group of images, they use PCA to discover the inter-image semantic correlation, and we use PCA within a single image to find the dominant object(s) of the scene.

Given features $\mathbf{x}$ of the image, $\mathbf{x}\in\mathbb{R}^{c\times wh}$,
we define the mean and covariance matrix of the features as:
\begin{equation}
\label{eq:mean_cov_feat}
\begin{aligned}
&\overline{\mathbf{x}}=\frac{1}{K} \sum_{p,q} \mathbf{x}_{(p,q)},\\
&\operatorname{Cov}(\mathbf{x})=\frac{1}{K}\sum_{p, q}\left(\mathbf{x}_{(p, q)}-\overline{\mathbf{x}}\right)\left(\mathbf{x}_{(p, q)}-\overline{\mathbf{x}}\right)^{\top},
\end{aligned}
\end{equation}
where $K=h\times w$, and
$\mathbf{x}_{(p,q)}$ indicates the feature vector with position coordinate $(p,q)$. 
We perform PCA on the covariance matrix $\operatorname{Cov}(\mathbf{x})$. 
PCA extracts the principal direction corresponding to the 
principal component of the features with the largest variance~\cite{pca1901}. Furthermore, as the object-centric representation has been generated by our semantic-guided learning network, we assume that the objects are the most distinctive component in the image representation.
In this case, the region that has a high correlation with the principal direction is regarded as the object region.
For each pixel, the projection map is calculated by:
\begin{equation}
\label{eq:pca_object_region}
\mathbf{m}=\xi_1^{\top}\left(\mathbf{x}-\overline{\mathbf{x}}\right),
\end{equation}
where $\xi_1\in\mathbb{R}^{c\times 1}$ represents the eigenvector corresponding to the maximal eigenvalue. 
As shown in Fig.~\ref{fig:pca_eigenvectors}, the projection map of the $1^{st}$ eigenvector ($1^{st}$ eigvec) with the largest eigenvalue can highlight the object, while the $2^{nd}$ and the $3^{rd}$ pay more attention to the background. Therefore, we use the projection map of the first eigenvector as our object discovery map.
The final binary segmentation map is produced by thresholding the projection map with a threshold of 0.5, which is the median value of the output map.

\section{Experimental Results}
\label{sec:experiments}
We evaluate our method in class-agnostic unsupervised object discovery in two forms: 1) Object Detection task to predict the bounding box of the object, and 2) Object Segmentation task to predict a binary segmentation map, including single-category object segmentation and generic-category salient object detection. Following the conventional practice from~\cite{wang2022self}, we also evaluate our method on video object detection.


\subsection{Setting}

\noindent\textbf{Evaluation Metrics:}
{\em Object Segmentation Metrics:} 1) $F$-measure: $F_\beta = \frac{(1+\beta^2) Precision\times Recall}{\beta^2 Precision + Recall}$, where $F^{\max}_{\beta}$ is reported as the maximal value of $F_\beta$ with 255 uniformly distributed binarization thresholds; 2) IoU (Intersection over Union) computes the overlapping between the ground-truth segmentation mask and binary predicted mask; 3) Accuracy measures the percentage of the correctly predicted pixels; 4) Jaccard index indicates the average IoU of the dataset. The threshold to generate binary masks for IoU and Accuracy is set to 0.5; 
{\em Object Detection Metrics:} 1) CorLoc is employed to measure the percentage of an object whose IOU with any ground truth bounding box in an image is larger than 0.5; 2) AP@50 is the average precision with 50$\%$ as the threshold for IoU, which
is used for multiple object detection.

\noindent\textbf{Implementation Details:} We select DINO-ViT-Small \cite{caron2021emerging} as the self-supervised transformer backbone, with patch size 16. During training, all images are resized to $224\times 224$. 
The base learning rate is set to be 7.5e-3 and updated by cosine rate decay. Maximum epoch is set to
200.
It takes an hour with batch size 48 on 6 NVIDIA GeForce RTX 3080 GPUs. The inference time for each image with resolution $480\times 480$ is around 
0.221s. Further, our model brings 2M extra parameters to our DINO-ViT-S/16 backbone.

\begin{table}[t!]
    \centering
    \footnotesize
    \renewcommand{\arraystretch}{1.2}
    \renewcommand{\tabcolsep}{1.9mm}
    \caption{Experimental results on unsupervised object discovery. $\ddag$ indicates the feature model pre-trained on ImageNet with class label (CorLoc).
    }
    \begin{tabular}{l|l|ccc}
    \toprule
    Methods & Feature & VOC07 & VOC12 & COCO20K \\ \hline
    Kim \etal \cite{kim2009unsupervised}& Hand-crafted & 43.9 & 46.4 & 35.1 \\
    DDT+\cite{wei2019unsupervised}&VGG19{$^{\ddag}$}& 50.2 & 53.1 & 38.2 \\
    rOSD \cite{unsup_multi_obj_dis}&VGG19{$^{\ddag}$}& 54.5 & 55.3 & 48.5 \\
    LOD \cite{large_unsup_obj_dis}&VGG16{$^{\ddag}$}& 53.6 & 55.1 & 48.5 \\
    Freesolo\cite{Freesolo2022} & ResNet50&44.0 & 49.7 & 35.2\\
    DeepSpectral\cite{Deep_Spectral_Methods_Kyriazi_2022_CVPR} & DINO-ViT-B/8 &62.7 &  66.4 & 52.2 \\
    DINO-Seg \cite{caron2021emerging}& DINO-ViT-S/16 & 45.8 & 46.2 & 42.1 \\
    LOST\cite{self_sup_loc_obj_no_labels}& DINO-ViT-S/16 & 61.9 & 64.0 & 50.7\\
    TokenCut\cite{wang2022self} & DINO-ViT-S/16 & \uline{68.8} & \bf{72.1} & \uline{58.8} \\
    \hline
    {\bf Ours} &DINO-ViT-S/16 & {\bf 70.6} & {\bf 72.1} & {\bf 63.5}\\
    \bottomrule
    \end{tabular}
    \label{tab:experiments_unsup_obj_discovery}
\end{table}

\begin{table*}[t!]
    \centering
    \footnotesize
    \renewcommand{\arraystretch}{1.2}
    \renewcommand{\tabcolsep}{1.1mm}
    \caption{
    Experiments on unsupervised object segmentation.
    }
    \begin{tabular}{l|l|ccc|ccc|ccc|ccc}
    \toprule
    & &\multicolumn{3}{c|}{CUB \cite{lin2014microsoft}}&\multicolumn{3}{c|}{ECSSD \cite{shi2015hierarchical}}&\multicolumn{3}{c|}{DUTS \cite{imagesaliency}}&\multicolumn{3}{c}{DUT-OMRON \cite{Manifold-Ranking:CVPR-2013}} \\
    Method & Pretraining & $ F_\beta^{\max}\uparrow$&IoU$\uparrow$&Acc.$\uparrow$& $ F_\beta^{\max}\uparrow$&IoU$\uparrow$&Acc.$\uparrow$& 
    $ F_\beta^{\max}\uparrow$&IoU$\uparrow$&Acc.$\uparrow$&
    $ F_\beta^{\max}\uparrow$&IoU$\uparrow$&Acc.$\uparrow$  \\
    \hline
    ReDO\cite{chen2019unsupervised} & SAGAN\cite{zhang2019self}+PSPNet\cite{zhao2017pyramid} & 48.1 & 36.0 & 85.3 & - & - & - & - & - & - & - & - & -  \\
    DRC\cite{yu2021unsupervised} & EBM \cite{pang2020learning} &50.2 & 39.9 & 85.8 & - & - & - & - & - & - & - & - & -  \\
    BigBiGAN\cite{object_seg_without_label_icml} & BigBiGAN\cite{donahue2019large} & 79.4 & 68.3 & 93.0  & 78.2 & 67.2 & 89.9 & 60.8 & 49.8 & 87.8 & 54.9 & 45.3 & 85.6 \\
    E-BigBiGAN\cite{object_seg_without_label_icml} & BigBiGAN\cite{donahue2019large} & \uline{83.4} & 71.0 & 94.0  & 79.7 & 68.4 & 90.6 & 62.4 & 51.1 & 88.2 & 56.3 & 46.4 & 86.0  \\
    FindGAN \cite{melas2021finding} & BigBiGAN\cite{donahue2019large} & 78.3 & 66.4 & 92.1  & \uline{80.6} & \uline{71.3} & 91.5 & 61.4 & 52.8 & 89.3 & 58.3 & 50.9 & \uline{88.3}  \\
    \hline
    LOST\cite{self_sup_loc_obj_no_labels} & DINO-ViT-S/16\cite{caron2021emerging} & 78.9 & 68.8 & 95.2 &  75.8 & 65.4 & 89.5 & 61.1 & 51.8 & 87.1 & 47.3 & 41.0 & 79.7 \\DeepSpectral\cite{Deep_Spectral_Methods_Kyriazi_2022_CVPR} & DINO-ViT-S/16\cite{caron2021emerging} & 82.9 & 66.7 & 94.1 & 78.5 & 64.5 & 86.4 & 62.1 & 47.1 & 84.1& 55.3 & 42.8 & 80.8 \\
    TokenCut\cite{wang2022self} & DINO-ViT-S/16\cite{caron2021emerging} & 82.1 & \uline{74.8} & \uline{96.4} & 80.3 & 71.2 & \uline{91.8} & \uline{67.2} & \uline{57.6} & \uline{90.3} & \uline{60.0} & \uline{53.3} & 88.0\\
    {\bf Ours} & DINO-ViT-S/16\cite{caron2021emerging} & {\bf 87.9} & {\bf 77.8} & {\bf 96.8} & {\bf 85.4} & {\bf 72.7} & {\bf 92.2} & {\bf 73.1} & {\bf 59.9} & {\bf 91.7} & {\bf 64.4} & {\bf 53.6} & {\bf 89.7}\\
    \hline
    \multicolumn{14}{c}{With Mask Refinement} \\
    \hline
    DeepSpectral\cite{Deep_Spectral_Methods_Kyriazi_2022_CVPR}+CRF & DINO-ViT-S/16\cite{caron2021emerging} & 84.3 & 77.6 & 96.6 & 80.5 & 73.3 & 89.1 & 64.4 & 56.7 & 87.1 & 56.7 & 51.4 & 83.8 \\
    LOST\cite{self_sup_loc_obj_no_labels}+BS\cite{barron2016fast}  & DINO-ViT-S/16\cite{caron2021emerging} & 84.1 & 74.2 & 96.5 &  83.7 & 72.3 & 91.6 & 69.7 & 57.2 & 88.7 & 57.8 & 48.9 & 81.8 \\
    TokenCut\cite{wang2022self}  + BS & DINO-ViT-S/16\cite{caron2021emerging} & \uline{87.1} & \uline{79.5} & {\bf 97.4} & \uline{87.4} & {\bf 77.2} & {\bf 93.4} & \uline{75.5} & \uline{62.4} & \uline{91.4} & {\bf 69.7} & {\bf 61.8} & \uline{89.7} \\
    
    {\bf Ours+BS}\cite{barron2016fast} & DINO-ViT-S/16\cite{caron2021emerging} & {\bf 89.3} & {\bf 79.7} & \uline{97.3} & {\bf 89.6} & \uline{74.2} & \uline{92.8} & {\bf 76.4} & {\bf 63.0} & {\bf 92.5} & \uline{68.3} & \uline{58.5} & {\bf 90.9}  \\
   \bottomrule
  \end{tabular}
  \label{tab:experiments_unsup_obj_seg}
\end{table*}

\subsection{Performance Comparison}
\label{sec:performance_comparision}

{\bf Object Detection:} We compare our proposed method with the state-of-the-art object detection methods \cite{caron2021emerging,self_sup_loc_obj_no_labels, Deep_Spectral_Methods_Kyriazi_2022_CVPR, wang2022self, large_unsup_obj_dis,unsup_multi_obj_dis,wei2019unsupervised,kim2009unsupervised, Freesolo2022}, and show CorLoc of the related models
in Table~\ref{tab:experiments_unsup_obj_discovery}. To train our model (\enquote{Ours}), we fine-tune DINO with RGB images from the DUTS \cite{imagesaliency} training dataset.
Except Freesolo~\cite{Freesolo2022}, the other compared methods in Table~\ref{tab:experiments_unsup_obj_discovery} are achieved directly on the features extracted from a fully/self-supervised backbone network pre-trained on ImageNet~\cite{russakovsky2015imagenet}, while Freesolo~\cite{Freesolo2022} is trained on COCO \texttt{train2017} and COCO \texttt{unlabeled2017}~\cite{lin2014microsoft} with a total of 241k training images. 
Since multiple objects are highlighted by our method, 
we generate the bounding box from the segmentation map by drawing the external rectangles containing both the whole foreground region and its connected components, where extremely small components are deleted. Specifically,
the steps of the bounding box generation process are:
\begin{itemize}
    \item The foreground region is partitioned by the 8-neighbour connected component analysis.
    \item Small connected components with area smaller than $0.25\%$ of the image are deleted.
    \item The components whose IoU with the complete foreground region is larger than 0.7 are deleted.
    \item The external rectangles for the remaining connected components and the whole foreground regions are drawn as the bounding boxes.
\end{itemize}
Differently, previous methods~\cite{self_sup_loc_obj_no_labels, wei2019unsupervised, unsup_multi_obj_dis} generate bounding boxes from the segmentation map by keeping the maximal connected component and filtering out other regions, leading to extensive
false negatives. 

According to Table~\ref{tab:experiments_unsup_obj_discovery}, our method exceeds the performance of other methods in terms of CorLoc score. From Table~\ref{tab:experiments_unsup_obj_discovery}, we can also infer that, compared with the methods based on hand-crafted features and supervised features, the methods based on self-supervised image representation, \ie~LOST~\cite{self_sup_loc_obj_no_labels}, TokenCut~\cite{wang2022self}, DeepSpectral~\cite{Deep_Spectral_Methods_Kyriazi_2022_CVPR} and our method achieves better results. However, due to the limited representation ability of the vanilla DINO feature, LOST~\cite{self_sup_loc_obj_no_labels}, TokenCut~\cite{wang2022self}, DeepSpectral~\cite{Deep_Spectral_Methods_Kyriazi_2022_CVPR} have inferior performance to our method. Specifically, we improve SOTA by $1.8\%$ and $5.7\%$ on VOC2007 and COCO20K and achieve comparable value with TokenCut on VOC2012.

In Fig.~\ref{fig:obj_detect_comparison}, we also show the visualization results of our method and three existing techniques.
It illustrates that the bounding box results of our method enclose more complete regions of the object than other approaches and contain less
background. In addition, as shown in the bottom row of Fig.~\ref{fig:obj_detect_comparison}, the proposed method works
better in
discovering multiple objects in the image compared with others. However, a falsely activated bounding box including all objects is also generated. Future work should be developed to address this issue.

\begin{table}[h]
    \centering
    \footnotesize
    \renewcommand{\arraystretch}{1.2}
    \renewcommand{\tabcolsep}{2.2mm}
    \caption{Performance of class-agnostic multiple object detection (CAD) with pseudo bounding box (AP@0.5).}
    \begin{tabular}{l|cccc}
    \toprule
    Method & VOC07 & VOC12 & COCO20K  \\
    \hline
    LOD~\cite{large_unsup_obj_dis}
    & 22.7 & 28.4 & 8.8 \\
    LOST~\cite{self_sup_loc_obj_no_labels}
    & \uline{29.0} & 33.5 & 9.9 \\
    TokenCut~\cite{wang2022self}
    & 26.2 & 35.0 & 10.5 \\
    UMOD~\cite{kara2023muod}
    & 27.9 & \uline{36.2} & {\bf 13.8}\\
    \hline
    {\bf Ours
    } & {\bf 30.5} & {\bf 36.8} &  \uline{13.6} \\
   \bottomrule
  \end{tabular}
  \label{tab:multiobject_detection}
\end{table}

\textbf{Class-Agnostic Multiple Object Detection:} To be more specific on how our model performs on multiple objects scenarios, we follow
the previous techniques~\cite{self_sup_loc_obj_no_labels} to report the Class-Agnostic multiple object Detection (CAD) 
results in Table~\ref{tab:multiobject_detection}, where the models are trained on VOC07 \texttt{trainval}~\cite{everingham2010pascal}, VOC12 \texttt{trainval}~\cite{everingham2011pascal}, COCO20K dataset~\cite{unsup_multi_obj_dis} and the testing is performed on VOC07 \texttt{test}, VOC12 \texttt{trainval} and COCO20K, respectively. Note that we report performance of the related models based on the provided numbers for multiple object detection.


To obtain results for multiple object detection, one can directly generate multiple bounding boxes from the segmentation map
as discussed in the previous section. However, the generated bounding boxes are too noisy to be directly used for multi-object detection performance evaluation.
Firstly, the connected instances will be enclosed in the same bounding box, leading to extensive false positive. Secondly, the object that is occluded by other objects will be split into multiple bounding boxes, which is the main cause of false negative. In addition, the CorLoc metric is insensitive to multiple objects and redundant objects. To mitigate the undesirable impact of noisy bounding boxes and better demonstrate the efficacy of our method on multiple object detection, the conventional practice is to train the class-agnostic Faster R-CNN model~\cite{girshick2015fastrcnn}, in which the object discovery bounding boxes are employed as the pseudo label for supervised learning, and performance in Table~\ref{tab:multiobject_detection} is obtained with extra training of a class-agnostic Faster R-CNN model~\cite{girshick2015fastrcnn}.
As illustrated in Table~\ref{tab:multiobject_detection}, performance of our model is comparable to the state-of-the-art multi-object detection model, namely UMOD~\cite{kara2023muod}. 
Differently from our solution, UMOD~\cite{kara2023muod} is also based on DINO~\cite{caron2021emerging} but has a more complex pipeline, including clustering of superpixels, merging the similar regions, training the foreground-background classifier, denoising the coarse masks and training the class-agnostic Fast R-CNN.

\begin{figure*}[tp]
   \begin{center}
   \begin{tabular}{{c@{ } c@{ } c@{ } c@{ } c@{ } c@{ } c@{ } c@{ }}}
    {\includegraphics[width=0.115\linewidth]{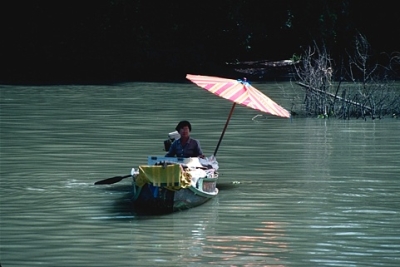}}&
    {\includegraphics[width=0.115\linewidth]{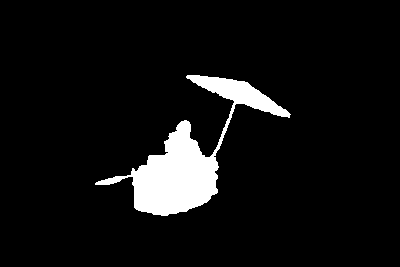}}&
    {\includegraphics[width=0.115\linewidth]{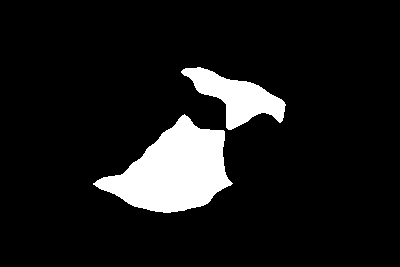}}&
    {\includegraphics[width=0.115\linewidth]{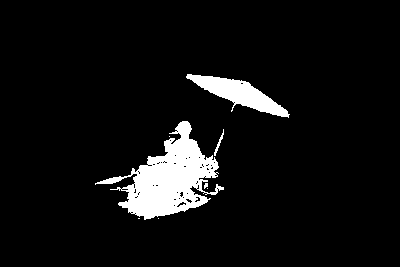}}&
   {\includegraphics[width=0.115\linewidth]{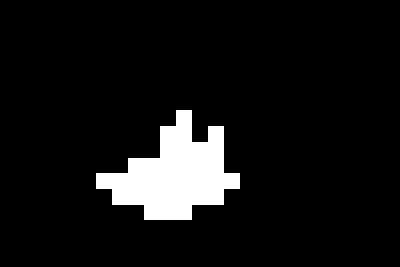}}&
    {\includegraphics[width=0.115\linewidth]{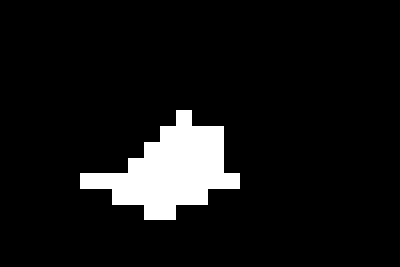}}&
    {\includegraphics[width=0.115\linewidth]{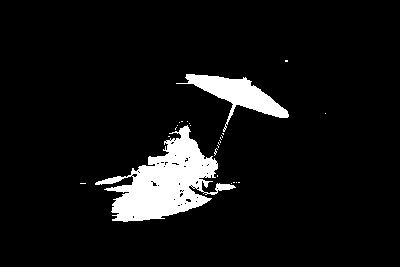}}&
    {\includegraphics[width=0.115\linewidth]{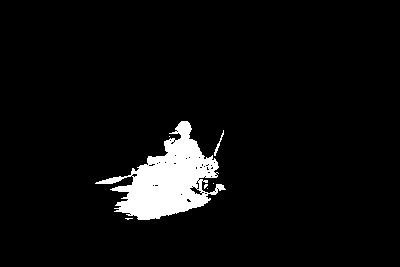}}\\
    {\includegraphics[width=0.115\linewidth]{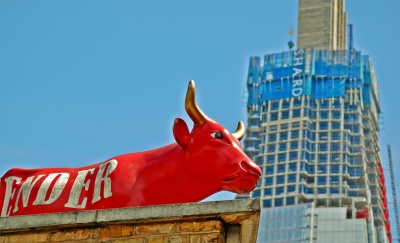}}&
    {\includegraphics[width=0.115\linewidth]{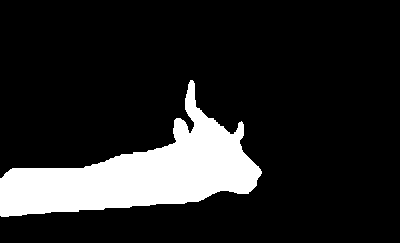}}&
    {\includegraphics[width=0.115\linewidth]{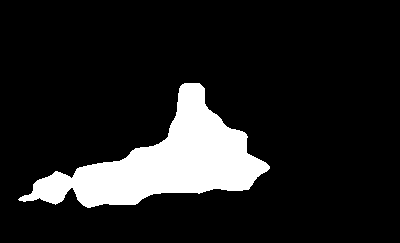}}&
    {\includegraphics[width=0.115\linewidth]{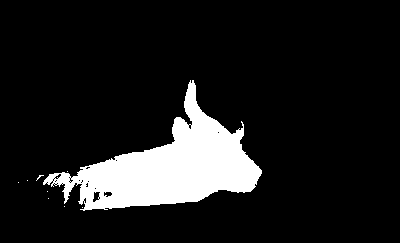}}&
   {\includegraphics[width=0.115\linewidth]{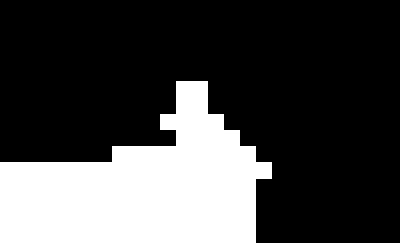}}&
    {\includegraphics[width=0.115\linewidth]{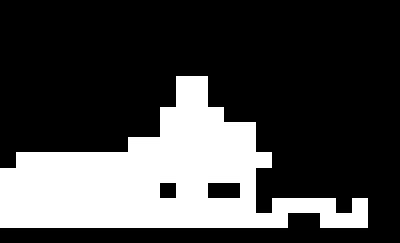}}&
    {\includegraphics[width=0.115\linewidth]{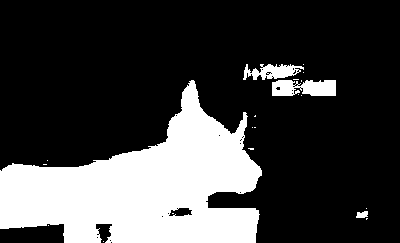}}&
    {\includegraphics[width=0.115\linewidth]{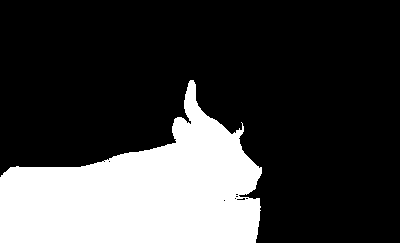}}\\
    {\includegraphics[width=0.115\linewidth]{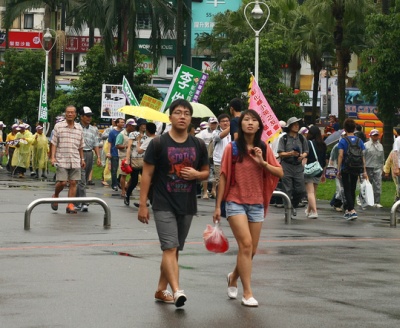}}&
    {\includegraphics[width=0.115\linewidth]{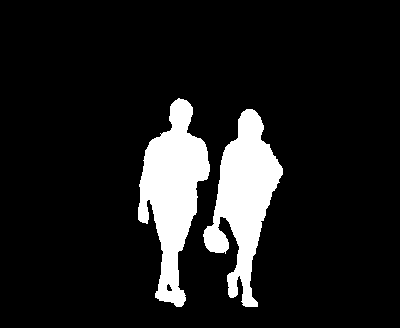}}&
    {\includegraphics[width=0.115\linewidth]{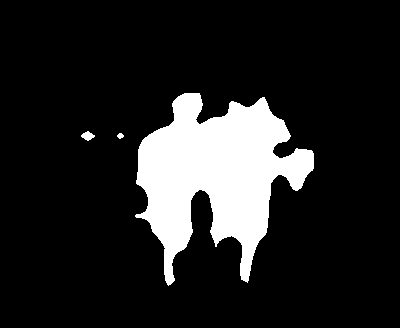}}&
    {\includegraphics[width=0.115\linewidth]{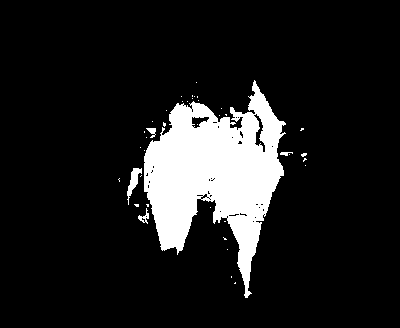}}&
   {\includegraphics[width=0.115\linewidth]{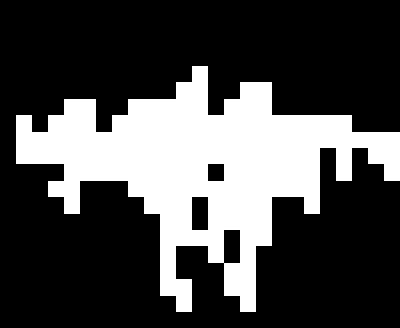}}&
    {\includegraphics[width=0.115\linewidth]{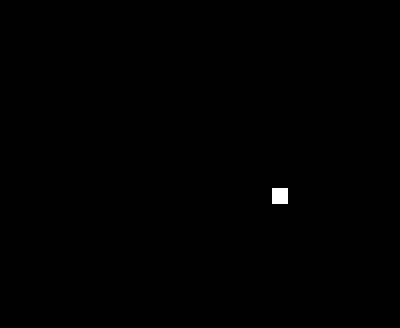}}&
    {\includegraphics[width=0.115\linewidth]{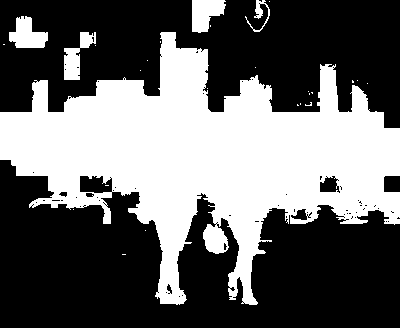}}&
    {\includegraphics[width=0.115\linewidth]{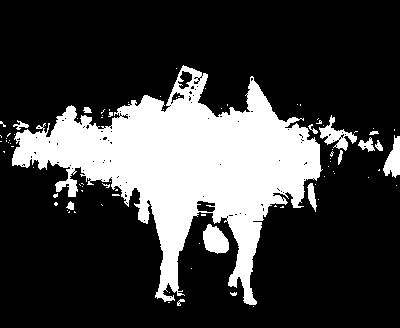}}\\
    \footnotesize{Image} &
     \footnotesize{GT} &\footnotesize{Ours}  &\footnotesize{Ours+BS} &\footnotesize{LOST\cite{self_sup_loc_obj_no_labels}} &\footnotesize{TokenCut \cite{wang2022self}}&\footnotesize{DeepSpectral\cite{Deep_Spectral_Methods_Kyriazi_2022_CVPR}}&\footnotesize{TokenCut \cite{wang2022self}+BS} \\
   \end{tabular}
   \end{center}
    \caption{Visual comparison of our method (\enquote{Ours}) with existing object segmentation models, where \enquote{BS} represents the bilateral solver (BS)~\cite{barron2016fast} as refinement. 
    } 
    \label{fig:obj_seg_comparison}
\end{figure*}

{\bf Object Segmentation:} We also compare the proposed method with the existing object segmentation methods\footnote{We test the missing metrics of the competing methods on our machine following their official implementation.} \cite{self_sup_loc_obj_no_labels, yu2021unsupervised,Deep_Spectral_Methods_Kyriazi_2022_CVPR, wang2022self,chen2019unsupervised,object_seg_without_label_icml,melas2021finding} as shown in Table~\ref{tab:experiments_unsup_obj_seg}, where
we train the model using images from the DUTS training dataset, and testing using CUB-200-2011 (CUB) \cite{lin2014microsoft} for single-category segmentation, DUTS \cite{imagesaliency}, ECSSD \cite{shi2015hierarchical} and DUT-OMRON \cite{Manifold-Ranking:CVPR-2013} datasets for multiple-category segmentation.
Among the generative solutions, ReDO~\cite{chen2019unsupervised} and DRC~\cite{yu2021unsupervised} train generative adversarial network (GAN)~\cite{gan_raw} or energy-based model (EBM)~\cite{pang2020learning} on the testing dataset of single object detection dataset such as CUB~\cite{lin2014microsoft} and Flowers~\cite{nilsback2007flowers} or simple multiple object detection dataset CLEVR~\cite{johnson2017clevr}.
BigBiGAN~\cite{object_seg_without_label_icml} and FindGAN~\cite{melas2021finding} train GAN model on ImageNet~\cite{russakovsky2015imagenet}. Other methods use the pre-trained backbone in DINO~\cite{caron2021emerging}.
We only report the results of \cite{chen2019unsupervised} and \cite{yu2021unsupervised} on the CUB bird dataset as their models were trained on a single-category dataset. 

Table~\ref{tab:experiments_unsup_obj_seg} demonstrates that the existing methods based on self-supervised image representation outperform both the generator methods \cite{yu2021unsupervised,chen2019unsupervised} and those generated from human-interpretable directions in the latent space of GANs \cite{object_seg_without_label_icml,melas2021finding} in both single-category dataset CUB and other generic-category salient object detection datasets.
Among all the related models, our method achieves the best result in all metrics before mask refinement.
We also show superior performance compared with the state-of-the-art in the majority of metrics after refinement with a bilateral solver \cite{barron2016fast}. Fig.~\ref{fig:obj_seg_comparison} illustrates the visualization results of our method in object segmentation. Compared with other algorithms, our method captures the prominent object even in complex surroundings. 

\begin{figure}[t!]
   \begin{center}
   \begin{tabular}{c@{ } c@{ } c@{ } c@{ } c@{ }}
    {\includegraphics[width=0.18\linewidth,height=0.16\linewidth]{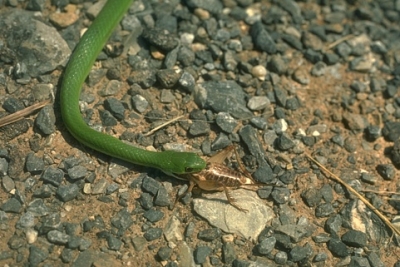}}&
    {\includegraphics[width=0.18\linewidth,height=0.16\linewidth]{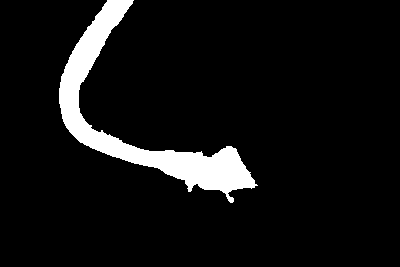}}&
    {\includegraphics[width=0.18\linewidth,height=0.16\linewidth]{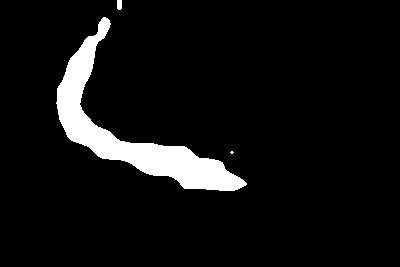}}&
    {\includegraphics[width=0.18\linewidth,height=0.16\linewidth]{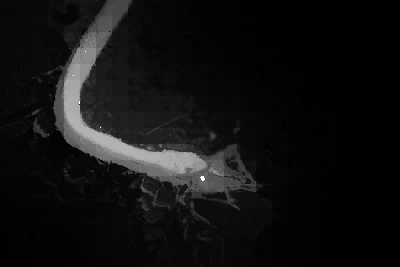}}&
    {\includegraphics[width=0.18\linewidth,height=0.16\linewidth]{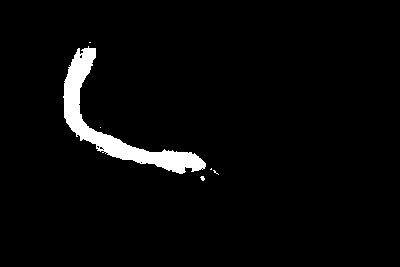}}\\
    {\includegraphics[width=0.18\linewidth,height=0.16\linewidth]{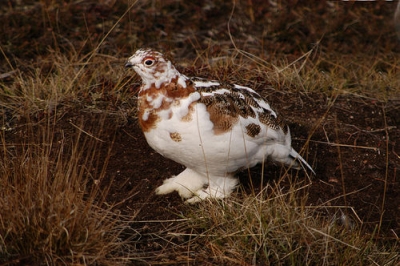}}&
    {\includegraphics[width=0.18\linewidth,height=0.16\linewidth]{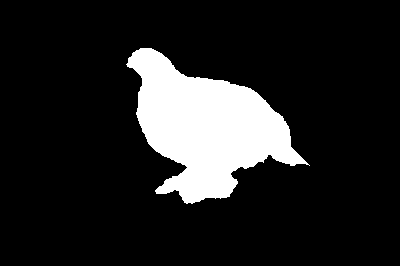}}&
    {\includegraphics[width=0.18\linewidth,height=0.16\linewidth]{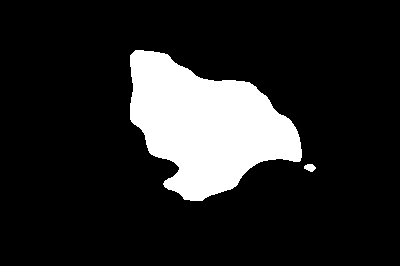}}&
    {\includegraphics[width=0.18\linewidth,height=0.16\linewidth]{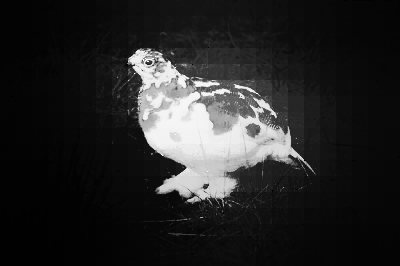}}&
    {\includegraphics[width=0.18\linewidth,height=0.16\linewidth]{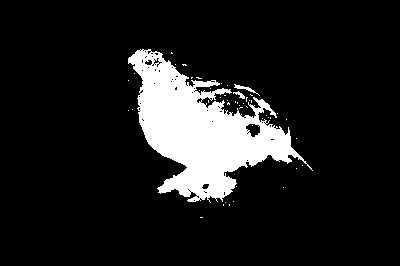}}\\
    {\includegraphics[width=0.18\linewidth,height=0.16\linewidth]{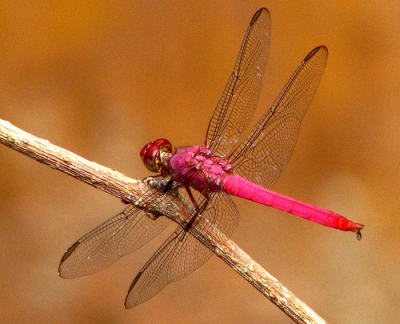}}&
    {\includegraphics[width=0.18\linewidth,height=0.16\linewidth]{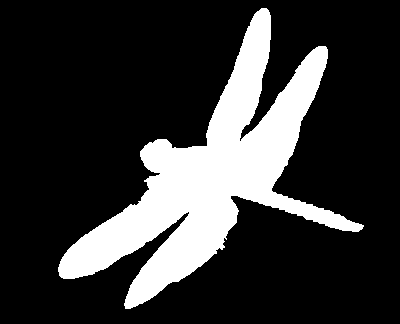}}&
    {\includegraphics[width=0.18\linewidth,height=0.16\linewidth]{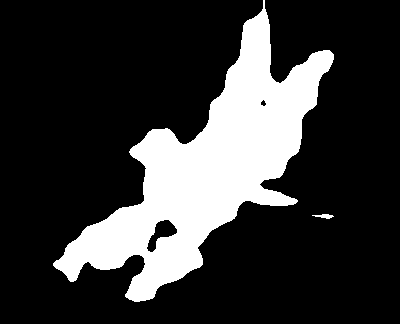}}&
    {\includegraphics[width=0.18\linewidth,height=0.16\linewidth]{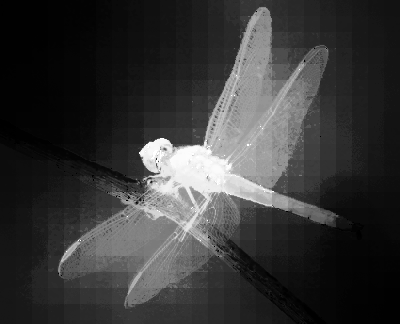}}&
    {\includegraphics[width=0.18\linewidth,height=0.16\linewidth]{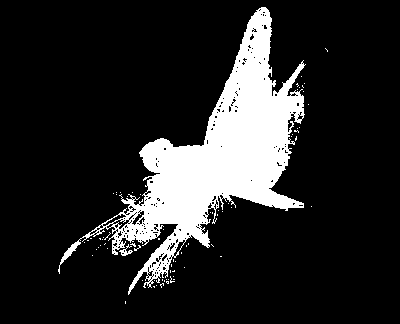}}\\
\footnotesize{Image} &\footnotesize{GT}  &\footnotesize{Ours} &\footnotesize{Ours+BS} &\footnotesize{BS+norm} \\
   \end{tabular}
   \end{center}
   \caption{Illustration of the bad effect of the bilateral solver on the segmentation performance. ``Ours+BS'' is the results after using the bilateral solver and ``BS+norm'' is the binary segmentation map produced by normalizing the ``Ours+BS'' results.}
   \label{fig:bad_bilateral_solver}
\end{figure}

\textit{Bilateral Solver \cite{barron2016fast} Analysis:} The bilateral solver~\cite{barron2016fast} is employed to post-process our coarse segmentation mask following TokenCut~\cite{wang2022self}. Although it is effective in most situations, according to the improvement shown in Table~\ref{tab:experiments_unsup_obj_seg}, there is still some negative influence on some images, as depicted in Fig.~\ref{fig:bad_bilateral_solver}. This is because bilateral solver segments object depending on the local structure of the image, and then the precision of the segmentation mask degrades when the local difference is large. Therefore, the unstable nature of the bilateral solver on segmentation is the main reason that our refined results is inferior to TokenCut~\cite{wang2022self} in some metrics. As a future work, we will be dedicated to improve the optimization objective in bilateral solver for more stable prediction refinement.

\textbf{Video Object Segmentation:} We also illustrate the generalization of our method by comparing our method with the training-free method TokenCut$\_$video~\cite{wang2022tokencut}, where our model is still trained with images from DUTS~\cite{imagesaliency} training dataset. We evaluate the model on the video dataset DAVIS2016~\cite{perazzi2016fdavis} (50 sequences, 3455 frames), SegTrackv2~\cite{li2013segtrackv2} (14 sequences, 1066 frames) and FBMS~\cite{ochs2013FBMS} (59 sequences, 7,306 frame, the annotation is provided every 20 frames).
To begin with, we obtain the RGB feature and Flow feature generated from RGB image and RAFT Flow image~\cite{teed2020raft}. Then we compute the covariance on these two features following Eq.~\eqref{eq:mean_cov_feat}, denoted as $\operatorname{Cov}$ and $\operatorname{CovF}$. Next, the first component is obtained by calculating the eigenvector of the matrix $\lambda_1\cdot\operatorname{Cov}+\lambda_2\cdot\operatorname{CovF}$, where $\lambda_1=0.5$, $\lambda_2=1.5$ in our setting. Since video segmentation should consider the inter-frame relationship, the eigenvector is computed by concatenating the features of multiple frames (20 frames in our experiment). From Table~\ref{tab:video_segmentation}, the performance is superior than TokenCut$\_$video on DAVIS~\cite{perazzi2016fdavis} and FBMS~\cite{ochs2013FBMS} dataset and a bit inferior on SegTrackv2~\cite{li2013segtrackv2} dataset. This is because SegTrackv2 is a small dataset that only contains 1066 frames and has more bias comparing with DAVIS and FBMS. In addition, the eigenvectors are calculated only once every 20 frames, and the speed can reach 20 FPS.

\begin{table}[h]
    \centering
    \footnotesize
    \renewcommand{\arraystretch}{1.2}
    \renewcommand{\tabcolsep}{1.8mm}
    \caption{Performance of video object segmentation (Jaccard Index).}
    \begin{tabular}{l|cccc}
    \toprule
    Method & DAVIS~\cite{perazzi2016fdavis} & FBMS~\cite{ochs2013FBMS} & SegTrackv2~\cite{li2013segtrackv2}  \\
    \hline
    TokenCut~\cite{wang2022self} & 64.3 & 60.2 & {\bf 59.6}\\
    {\bf Ours} & {\bf 69.0} & {\bf 60.4} & 57.6 \\
   \bottomrule
  \end{tabular}
  \label{tab:video_segmentation}
\end{table}

\begin{table*}[t!]
    \centering
    \footnotesize
    \renewcommand{\arraystretch}{1.2}
    \renewcommand{\tabcolsep}{2.5mm}
    \caption{
    Ablation study on the effectiveness of DINO feature and different loss functions of the proposed strategy.
    }
    \begin{tabular}{l|ccc|ccc|ccc|ccc}
    \toprule
    & \multicolumn{3}{c|}{CUB \cite{lin2014microsoft}}&\multicolumn{3}{c|}{ECSSD \cite{shi2015hierarchical}}&\multicolumn{3}{c|}{DUTS \cite{imagesaliency}}&\multicolumn{3}{c}{DUT-OMRON \cite{Manifold-Ranking:CVPR-2013}} \\
    Method  & $ F_\beta^{\max}\uparrow$&IoU$\uparrow$&Acc.$\uparrow$& $ F_\beta^{\max}\uparrow$&IoU$\uparrow$&Acc.$\uparrow$& 
    $ F_\beta^{\max}\uparrow$&IoU$\uparrow$&Acc.$\uparrow$&
    $ F_\beta^{\max}\uparrow$&IoU$\uparrow$&Acc.$\uparrow$  \\
    \hline
    vanilla DINO+PCA & 77.7 & 65.7 & 92.9 & 75.4 & 64.7 & 88.6 & 57.0  & 46.1 & 86.0 & 50.1 & 41.4  & 84.5 \\
    vanilla DINO+SC & 82.1 & 74.8 & 96.4 & 80.3 & 71.2 & 91.8 & 67.2 & 57.6 & 90.3 & 60.0 & 53.3 & 88.0\\
    finetuned DINO+PCA & 61.2 & 51.7 & 88.6 & 67.8 & 55.6 & 85.5 & 45.9 & 36.7 & 83.0 & 41.8 & 34.1 & 82.0 \\
    finetuned DINO+SC & 80.2 & 67.9 & 93.3 & 80.1 & 71.1 & 91.6  & 60.0 & 52.3 & 86.6 & 55.7 & 49.7 & 84.5 \\
    \hline
    NCE Loss & 81.0 & 70.8 & 94.5 & 83.1 & 69.7 & 91.2 & 63.6 & 52.4 & 88.3 & 56.7 & 47.4 & 86.6 \\
    Graph Loss & 85.3 & 74.2 & 95.7 & 82.5 & 70.9 & 91.4 & 67.0 & 54.7 & 89.2 & 58.2 & 48.3 & 86.8 \\
    w/o Align Loss & 85.8 & 74.4 & 95.7 & 82.6 & 71.0 & 91.3 & 67.5 & 55.4 & 89.4 & 59.3 & 49.6 & 87.3 \\
    w/o Graph Loss & 84.2 & 73.4 & 95.4  & 83.9 & 71.2 & 91.7 & 66.2 & 54.0 & 89.1 & 58.2 & 48.6  & 87.3 \\
    w/o NCE Loss & 87.5 & 76.0 & 96.4 & 82.9 & 71.1 & 91.5 & 69.4  & 56.5 & 90.3 & 60.7 & 50.2  & 88.2 \\
    {\bf Ours} & {\bf 87.9} & {\bf 77.8} & {\bf 96.8} & {\bf 85.4} & {\bf 72.7} & {\bf 92.2} & {\bf 73.1} & {\bf 59.9} & {\bf 91.7} & {\bf 64.4} & {\bf 53.6} & {\bf 89.7}\\
   \bottomrule
  \end{tabular}
  \label{tab:ablation_study_cod}
\end{table*}
\begin{table*}[h]
\setcounter{table}{5}
    \centering
    \footnotesize
    \renewcommand{\arraystretch}{1.2}
    \renewcommand{\tabcolsep}{2.8mm}
    \caption{Experiments of different self-supervised methods (row 1-4), deep unsupervised SOD model (row 5-6) and the comparison between spectral clustering based discovery methods and PCA-based discovery methods (row 7-10).
    }
    \begin{tabular}{l|l|ccc|ccc|ccc}
    \toprule
    & &\multicolumn{3}{c|}{ECSSD \cite{shi2015hierarchical}}&\multicolumn{3}{c|}{DUTS \cite{imagesaliency}}&\multicolumn{3}{c}{DUT-OMRON \cite{Manifold-Ranking:CVPR-2013}} \\
    Method & Pretraining & $ F_\beta^{\max}\uparrow$&IoU$\uparrow$&Acc.$\uparrow$& 
    $ F_\beta^{\max}\uparrow$&IoU$\uparrow$&Acc.$\uparrow$&
    $ F_\beta^{\max}\uparrow$&IoU$\uparrow$&Acc.$\uparrow$  \\
    \hline
    TokenCut\cite{wang2022self} & MAE-ViT-B/16\cite{mae_he_2022}& 72.0  & 65.2 & 87.6 & 53.9 & 47.7 & 83.7& 46.4 & 41.5 & 78.0 \\
    {\bf Ours} & MAE-ViT-B/16\cite{mae_he_2022} & {\bf 77.1} & {\bf 67.1} & {\bf 87.7} & {\bf 63.0} & {\bf 51.1} & {\bf 86.1} & {\bf 56.3} & {\bf 47.0} & {\bf 85.2}  \\
    \hline
    TokenCut\cite{wang2022self} & MocoV3-ViT-S/16\cite{mocov3_he_2021} & 79.0 & 69.0 & 90.2 & 61.9 & 53.5 & 87.6 & 49.8 &  43.2 & 81.7  \\
    {\bf Ours} & MocoV3-ViT-S/16\cite{mocov3_he_2021} & {\bf 84.2} & {\bf 73.5} & {\bf 92.1} & {\bf 70.3} & {\bf 58.1} & {\bf 90.6} & {\bf 62.9} & {\bf 52.5} & {\bf 89.1}  \\    
    \hline
    SelfMask\cite{shin2022unsupervised} & Maskformer\cite{maskformer} & {\bf 88.9}  &  78.1  &  94.4  &  75.0  &  62.6  &  92.3  &  68.0  &  58.2  &  90.1  \\
    {\bf Ours$\_$pseudoMask} & Maskformer\cite{maskformer} & 88.8 & {\bf 79.4}  & {\bf 94.5}  & {\bf 76.1}  & {\bf 63.4} & {\bf 93.4} & {\bf 71.3}  & {\bf 60.3}  & {\bf 92.5}   \\
    \hline
    DeepSpectral\cite{Deep_Spectral_Methods_Kyriazi_2022_CVPR} & DINO-ViT-S/16\cite{caron2021emerging}  & 78.5 & 64.5 & 86.4 & 62.1 & 47.1 & 84.1& 55.3 & 42.8 & 80.8 \\
    TokenCut\cite{wang2022self} & DINO-ViT-S/16\cite{caron2021emerging} & 80.3 & \uline{71.2} & \uline{91.8} & 67.2 & 57.6 & 90.3 & 60.0 & \uline{53.3} & 88.0\\
    Ours$\_$SC & DINO-ViT-S/16\cite{caron2021emerging} & \uline{83.0} & 66.0 & 91.0  &  \uline{72.5} & \uline{58.3} & {\bf 92.2} & \uline{63.8} & 47.0 & \uline{89.4} \\
    {\bf Ours$\_$PCA} & DINO-ViT-S/16\cite{caron2021emerging} & {\bf 85.4} & {\bf 72.7} & {\bf 92.2} & {\bf 73.1} & {\bf 59.9} & \uline{91.7} & {\bf 64.4} & {\bf 53.6} & {\bf 89.7}\\  
   \bottomrule
  \end{tabular}
  \label{tab:experiments_unsup_obj_seg_discussion}
\end{table*}

\subsection{Ablation Study}
We have also conducted extra experiments to comprehensively analyse different components of the proposed strategy, and show performance in
Table~\ref{tab:ablation_study_cod}. 
\enquote{Vanilla DINO} in Table~\ref{tab:ablation_study_cod} indicates that the object region is discovered by employing PCA or spectral clustering directly on the original DINO feature and \enquote{finetuned DINO} represents that the backbone is further trained using images of the DUTS training dataset~\cite{imagesaliency}, following the original loss in DINO~\cite{caron2021emerging}. Row 1-4 in Table~\ref{tab:ablation_study_cod} show that the finetuned DINO model has even worse performance in localizing objects than the original DINO model. Our method is then proposed to solve the issue that the representation is disturbed by complicated backgrounds when training on scene-centric images.

Then, we show the importance of the losses in our method. In Table~\ref{tab:ablation_study_cod}, \enquote{NCE Loss} and \enquote{Graph Loss}
denote that only NCE Loss defined in Eq.~\eqref{infornce_loss} or Graph Loss defined in Eq.~\eqref{graph_loss} is used in finetuning the vanilla DINO-pretrained backbone, respectively. We can see that their performance are both improved compared with \enquote{vanilla DINO+PCA}, but still inferior to \enquote{vanilla DINO+SC}. The reason for not using alignment loss in Eq.~\eqref{eq:alignment_loss} alone is that training the network with it alone cannot converge.
In addition, the performance drops if one of the losses in our semantic-guided representation learning method is deleted, which demonstrates that all of these losses are essential.

\subsection{Discussions}
\noindent\textbf{Using other self-supervised pre-training model:} The objective of our method is to make the model features concentrate on the object. Therefore, it could be generalized to other self-supervised learning methods. To evaluate its effectiveness, we replace the ViT model pretrained by DINO with MocoV3 \cite{mocov3_he_2021} and MAE \cite{mae_he_2022} pretraining model. As shown in Table~\ref{tab:experiments_unsup_obj_seg_discussion}, the performance is generally better when WCL and alignment loss are employed.



\noindent\textbf{Discovery map as pseudo mask for segmentation model:} In order to further improve the performance of segmentation, the discovery mask could be employed as the pseudo mask to train a segmentation model~\cite{Freesolo2022, shin2022unsupervised}. Therefore, we train the salient object detection model based on Maskformer~\cite{maskformer} by using our discovery mask as the pseudo mask. From Table~\ref{tab:experiments_unsup_obj_seg_discussion}, we can find that the performance is superior to the SOTA Selfmask~\cite{shin2022unsupervised} under most metrics, even though three types of pseudo labels are used in Selfmask. 


\begin{table}[h]
    \centering
    \footnotesize
    \renewcommand{\arraystretch}{1.2}
    \renewcommand{\tabcolsep}{3.8mm}
    \caption{Performance on large dataset C120K. ``V07$\_$TV'' and ``V07$\_$TE'' are the ``trainval'' and ``test'' subset of VOC2007. }
    \begin{tabular}{l|l|ccc}
    \toprule
    Method & Pretraining & CorLoc  \\
    \hline
    LOD\cite{large_unsup_obj_dis} & VGG16 & 48.6 \\
    LOD\cite{large_unsup_obj_dis} & VGG16-Self & 42.4 \\
    TokenCut\cite{wang2022self} & DINO-ViT-S/16 & 58.5  \\
    {\bf Ours} & DINO-ViT-S/16 & {\bf 63.3} \\
   \bottomrule
  \end{tabular}
  \label{tab:large_dataset}
\end{table}

{\noindent\bf Scalability on large dataset: }
We apply our method to C120K~\cite{large_unsup_obj_dis} dataset to verify the scalability of our method.
Images in C120K are selected from the training and validation sets of the COCO2014 dataset~\cite{lin2014microsoft}, but those containing only the {\it crowd} object are deleted. As shown in Table~\ref{tab:large_dataset}, both TokenCut~\cite{wang2022self} and our method can scale well on the C120K dataset, while we achieve superior 
performance.

\begin{figure}[t!]
   \begin{center}
   \begin{tabular}{c@{ } c@{ } c@{ } c@{ }}
   \rotatebox{90}{{~~~~~~GT}}
    {\includegraphics[width=0.22\linewidth,height=0.18\linewidth]{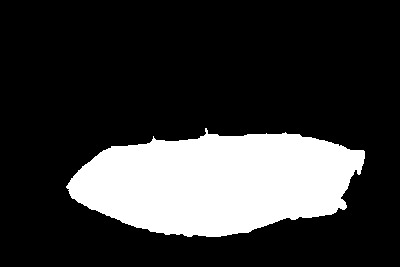}}&
    {\includegraphics[width=0.22\linewidth,height=0.18\linewidth]{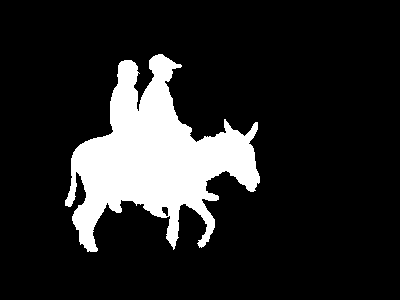}}&
    {\includegraphics[width=0.22\linewidth,height=0.18\linewidth]{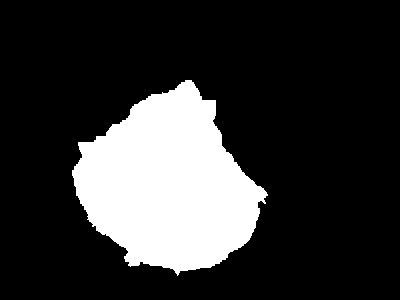}}&
    {\includegraphics[width=0.22\linewidth,height=0.18\linewidth]{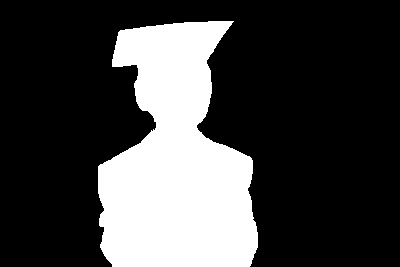}}\\
    \rotatebox{90}{{~~~~~~SC}}
    {\includegraphics[width=0.22\linewidth,height=0.18\linewidth]{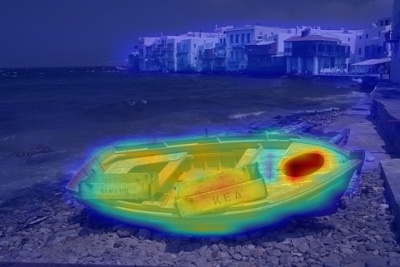}}&
    {\includegraphics[width=0.22\linewidth,height=0.18\linewidth]{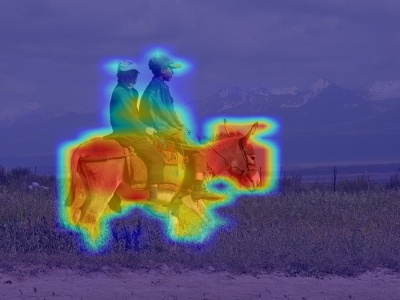}}&
    {\includegraphics[width=0.22\linewidth,height=0.18\linewidth]{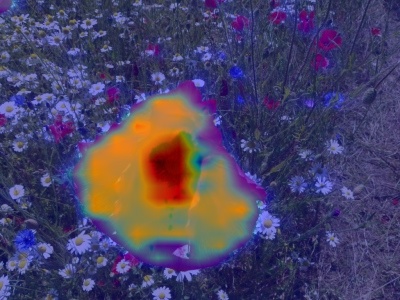}}&
    {\includegraphics[width=0.22\linewidth,height=0.18\linewidth]{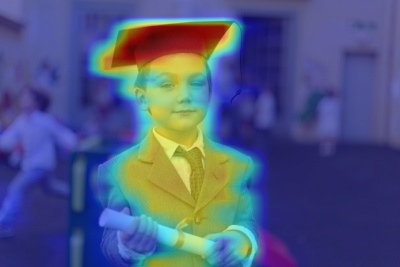}}\\
    \rotatebox{90}{{~~~~~~PCA}}
    {\includegraphics[width=0.22\linewidth,height=0.18\linewidth]{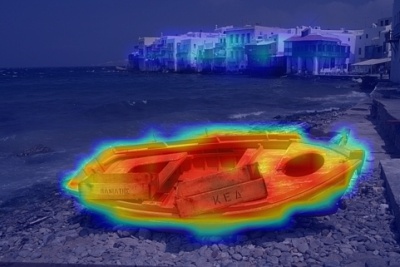}}&
    {\includegraphics[width=0.22\linewidth,height=0.18\linewidth]{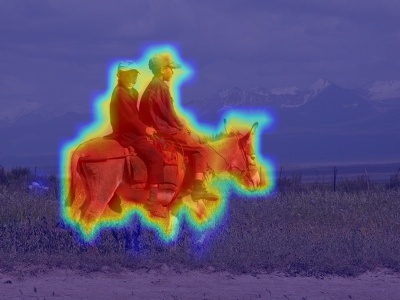}}&
    {\includegraphics[width=0.22\linewidth,height=0.18\linewidth]{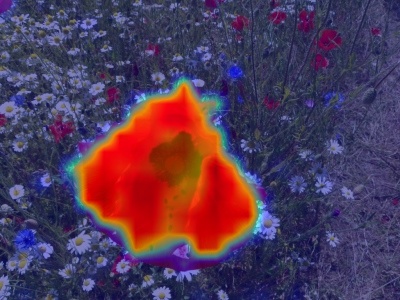}}&
    {\includegraphics[width=0.22\linewidth,height=0.18\linewidth]{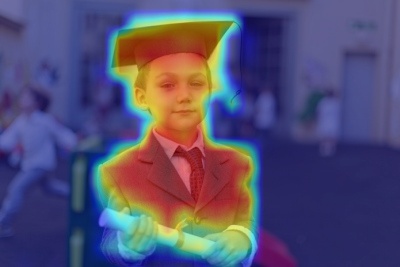}}\\
\footnotesize{(a)} &\footnotesize{(b)}  &\footnotesize{(c)} &\footnotesize{(d)} \\
   \end{tabular}
   \end{center}
   \caption{Comparison between spectral clustering (SC) and PCA on ViT features trained by our semantic-guided self-supervised learning.}
   \label{fig:pca_spClus}
\end{figure}

{\noindent\bf Spectral clustering v.s. PCA:} In early works, such as TokenCut~\cite{wang2022self} and DeepSpectral~\cite{Deep_Spectral_Methods_Kyriazi_2022_CVPR}, spectral clustering (SC) has shown its advantage for object discovery. Although both SC and PCA are dimension reduction methods, experimental results show that our proposed PCA-based discovery method outperforms
the SC-based discovery method. To verify effectiveness of PCA within our setting, we replace it with
spectral clustering following the same procedure as it has been used in
TokenCut~\cite{wang2022self} and report the performance in Table~\ref{tab:experiments_unsup_obj_seg_discussion} as Ours$\_$SC. From the results, although Ours$\_$SC has better values in most metrics, our PCA-based method (Ours$\_$PCA) outperforms the spectral clustering version
(Ours$\_$SC) generally. The reason is that our weakly-supervised contrastive learning (WCL) loss and alignment loss enable the network to focus more on the discriminative
features of the object. With PCA on top of the WCL, our method can capture the features of the object more completely, while
spectral clustering localize mainly on the
local regions of the object, representing the discriminative region (see Fig.~\ref{fig:pca_spClus}). 

\section{Failure cases}

\subsection{Object detection}
\begin{figure}[t!]
   \begin{center}
   \begin{tabular}{c@{ } c@{ } c@{ } c@{ }}
   \rotatebox{90}{{~~~~~~GT}}
    {\includegraphics[width=0.22\linewidth,height=0.18\linewidth]{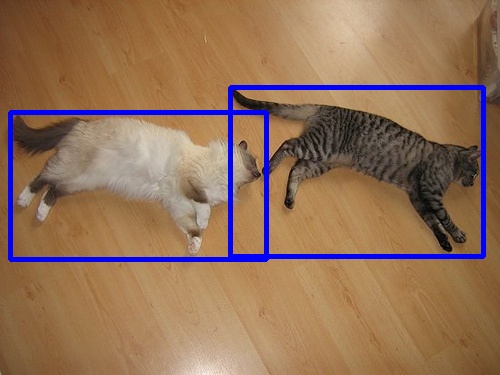}}&
    {\includegraphics[width=0.22\linewidth,height=0.18\linewidth]{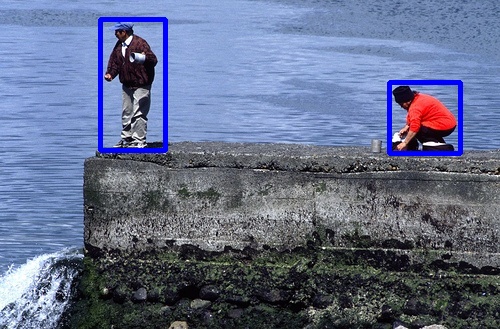}}&
    {\includegraphics[width=0.22\linewidth,height=0.18\linewidth]{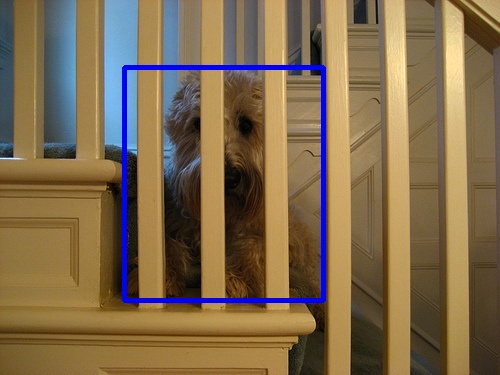}}&
    {\includegraphics[width=0.22\linewidth,height=0.18\linewidth]{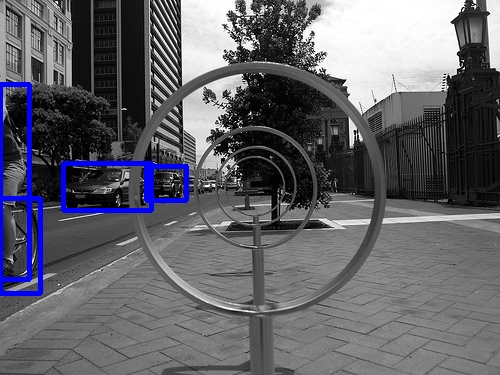}}\\
    \rotatebox{90}{{~~~~~Ours}}
    {\includegraphics[width=0.22\linewidth,height=0.18\linewidth]{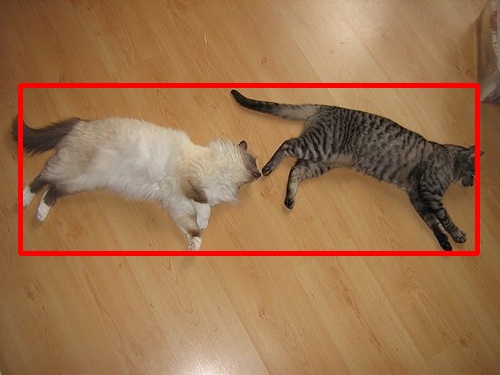}}&
    {\includegraphics[width=0.22\linewidth,height=0.18\linewidth]{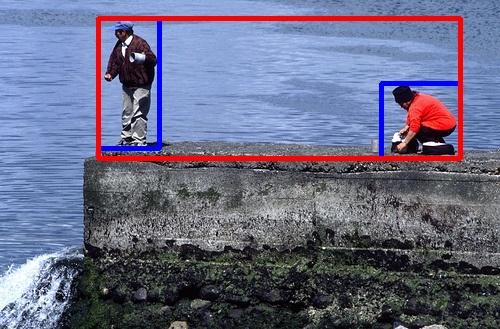}}&
    {\includegraphics[width=0.22\linewidth,height=0.18\linewidth]{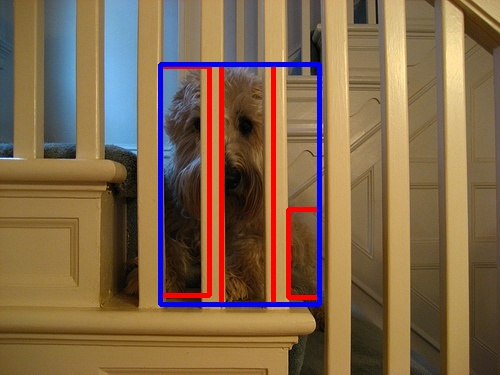}}&
    {\includegraphics[width=0.22\linewidth,height=0.18\linewidth]{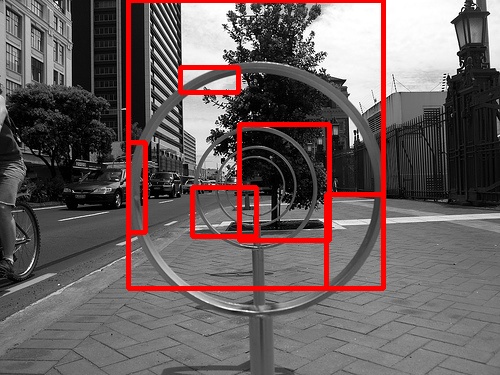}}\\
    \rotatebox{90}{{~~Heatmap}}
    {\includegraphics[width=0.22\linewidth,height=0.18\linewidth]{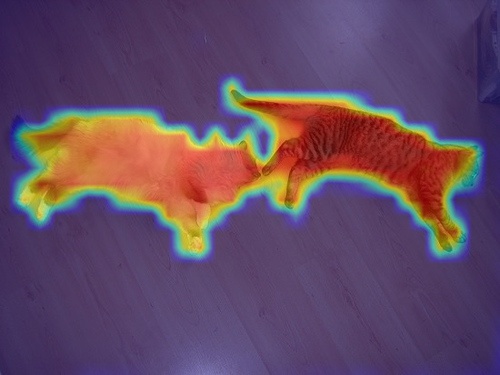}}&
    {\includegraphics[width=0.22\linewidth,height=0.18\linewidth]{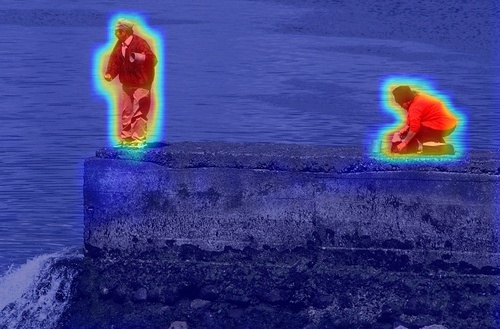}}&
    {\includegraphics[width=0.22\linewidth,height=0.18\linewidth]{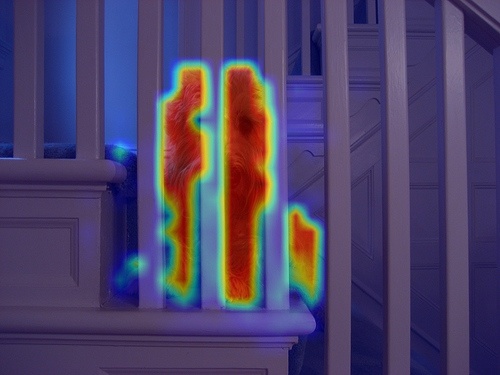}}&
    {\includegraphics[width=0.22\linewidth,height=0.18\linewidth]{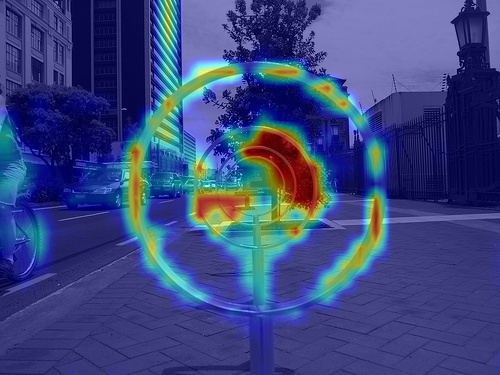}}\\
\footnotesize{(a)} &\footnotesize{(b)}  &\footnotesize{(c)} &\footnotesize{(d)} \\
   \end{tabular}
   \end{center}
   \caption{Visualization of failure cases for object detection. Bounding boxes with red color represent redundant or error results while blue bounding boxes indicate the correct ones.}
   \label{fig:fail_bbox}
\end{figure}

The failure of object detection is mainly attributed to the bounding box generation method and inaccurate segmentation results. Fig.~\ref{fig:fail_bbox} displays different situations where the failure cases happen, in which (a)-(c) are the cases caused by the bounding box generation method and (d) is the failure case generated by the error segmentation map. We can see that, a) the object detection cannot split the connected component region that contains multiple objects, b) the whole region bounding box is redundant if multiple objects are successfully captured by the separated bounding boxes, and c) the separated bounding boxes generated by different connected components are unwanted if the object is occluded by other objects. In addition, as shown in (d), our object discovery method fails to capture the objects that are less salient in the image.

\subsection{Object segmentation}
\begin{figure}[t!]
   \begin{center}
   \begin{tabular}{c@{ } c@{ } c@{ } c@{ }}
   \rotatebox{90}{{~~~~Image}}
    {\includegraphics[width=0.22\linewidth,height=0.18\linewidth]{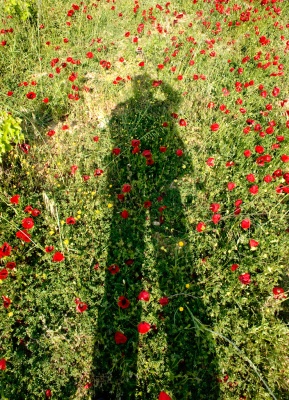}}&
    {\includegraphics[width=0.22\linewidth,height=0.18\linewidth]{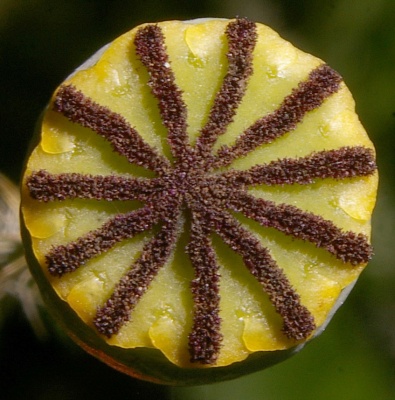}}&
    {\includegraphics[width=0.22\linewidth,height=0.18\linewidth]{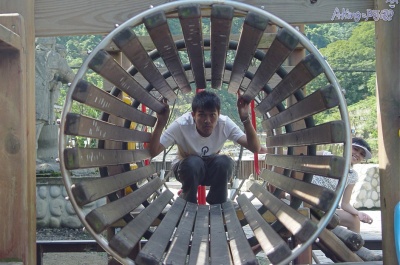}}&
    {\includegraphics[width=0.22\linewidth,height=0.18\linewidth]{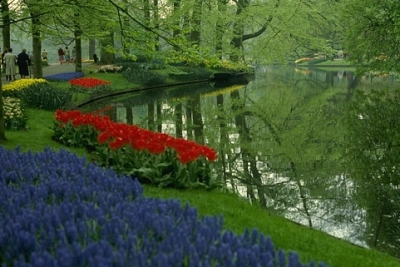}}\\
    \rotatebox{90}{{~~~~~GT}}
    {\includegraphics[width=0.22\linewidth,height=0.18\linewidth]{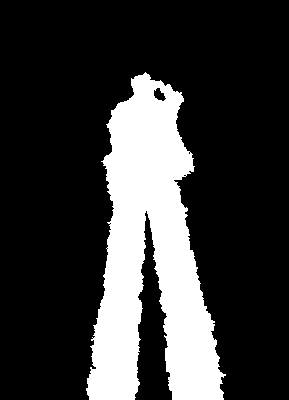}}&
    {\includegraphics[width=0.22\linewidth,height=0.18\linewidth]{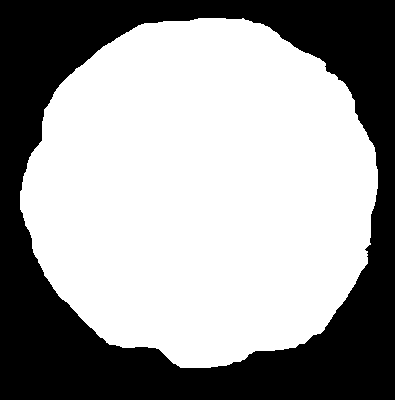}}&
    {\includegraphics[width=0.22\linewidth,height=0.18\linewidth]{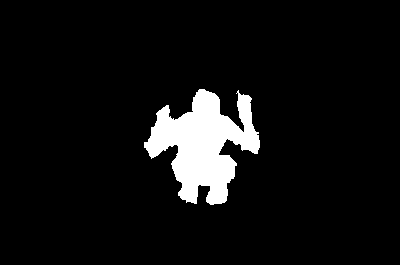}}&
    {\includegraphics[width=0.22\linewidth,height=0.18\linewidth]{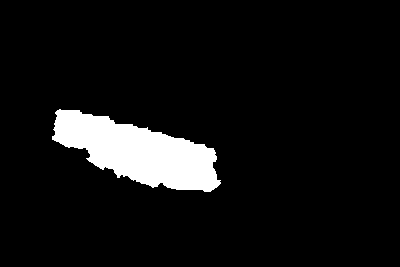}}\\
    \rotatebox{90}{{~~~~~Ours}}
    {\includegraphics[width=0.22\linewidth,height=0.18\linewidth]{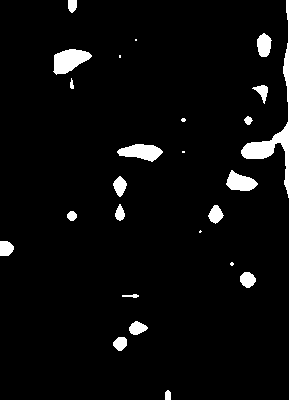}}&
    {\includegraphics[width=0.22\linewidth,height=0.18\linewidth]{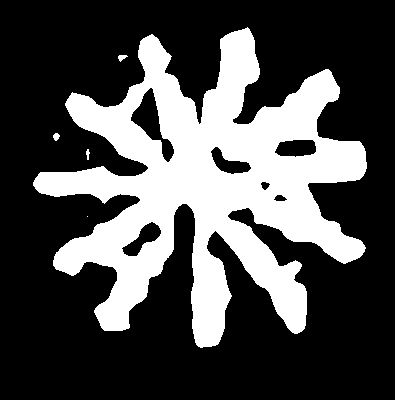}}&
    {\includegraphics[width=0.22\linewidth,height=0.18\linewidth]{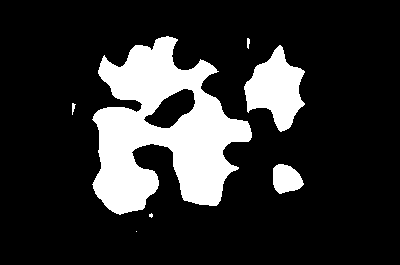}}&
    {\includegraphics[width=0.22\linewidth,height=0.18\linewidth]{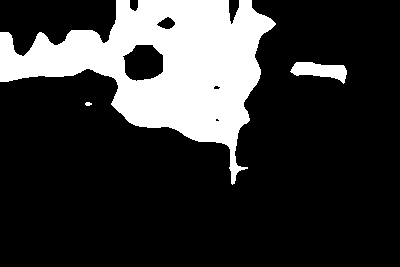}}\\
    \rotatebox{90}{{~~Heatmap}}
    {\includegraphics[width=0.22\linewidth,height=0.18\linewidth]{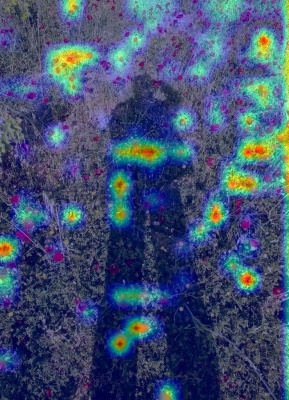}}&
    {\includegraphics[width=0.22\linewidth,height=0.18\linewidth]{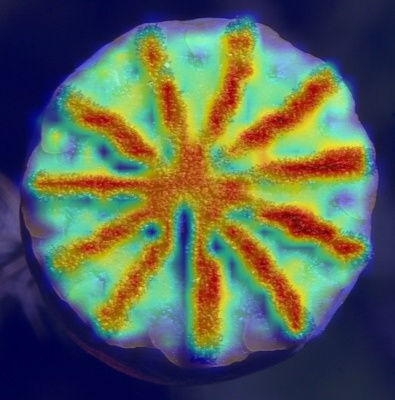}}&
    {\includegraphics[width=0.22\linewidth,height=0.18\linewidth]{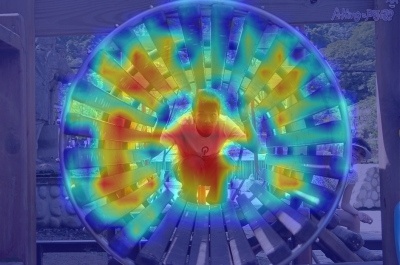}}&
    {\includegraphics[width=0.22\linewidth,height=0.18\linewidth]{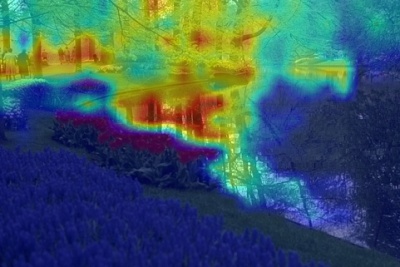}}\\
\footnotesize{(a)} &\footnotesize{(b)}  &\footnotesize{(c)} &\footnotesize{(d)} \\
   \end{tabular}
   \end{center}
   \caption{Visualization of failure cases for object segmentation.}
   \label{fig:fail_seg}
\end{figure}

Fig.~\ref{fig:fail_seg} illustrates some failure cases of object segmentation. In (a), the contour of the object has a human structure, but we fail to segment it because further details are required. In (b), although the salient parts of the object are highlighted in the discovery map, our method ignores the less salient regions that also belong to the object. If the background also contains some regions with an objectness property, some regions in the background will also be explored (in (c)). Conversely, if the object has less objectness, the object in (d) could not be discovered because it is sometimes a component of the background in other scene-centric images.

\section{Conclusion}
In this paper, we propose an unsupervised object discovery method based on semantic-guided self-supervised representation learning.
Our method uses the DINO transformer backbone as the feature encoder, benefiting from the large-scale object-centric pre-training. Then we finetune the DINO backbone on the DUTS training images using our semantic-guided representation learning method to extract the object-centric representation from scene-centric data. Finally, the object region is generated from the image representation by extracting the principal component via PCA. Experiments conducted on object detection, object segmentation and video object segmentation demonstrate the effectiveness of our method. As a self-supervised learning technique, similar to existing techniques, our method still has limitations in dealing with images of complex background. More investigation into robust unsupervised object detection will be conducted in the future.

\bibliographystyle{IEEEtran}
\bibliography{tip_reference}

\newpage

 




\vfill

\end{document}